\documentclass[sigconf]{acmart}
\usepackage[mathscr]{eucal}
\usepackage{amsmath,amsfonts,amsthm,bm}
\usepackage{enumitem}
\usepackage{tabularx}
\usepackage{multirow}
\usepackage{booktabs}
\usepackage{scalerel}
\usepackage{subcaption}    
\usepackage{amsmath}
\usepackage[multiple]{footmisc}
\usepackage{tikz}
\usepackage{pgfplots}
\usepackage{pifont}
\usepackage{algorithm} 
\usepackage{algpseudocode}
\usepackage{tabularx}
\usepackage{footnote}
\usepackage{balance}
\usepackage{graphicx}
\usepackage{nicefrac}
\usepackage{subcaption}
\usepackage{tikz}
\usepackage{textcomp}
\usepackage{makecell}
\usepackage{multirow}
\usetikzlibrary{patterns}

\makeatletter
\def\BibTeX{{\rm B\kern-.05em{\sc i\kern-.025em b}\kern-.08em
    T\kern-.1667em\lower.7ex\hbox{E}\kern-.125emX}}

\newcommand{\multiline}[1]{%
  \begin{tabularx}{\dimexpr\linewidth-\ALG@thistlm}[t]{@{}X@{}}
    #1
  \end{tabularx}
}

\pgfplotsset{compat=1.5.1}
\def\addlegendimage{\csname pgfplots@addlegendimage\endcsname}
\usetikzlibrary{patterns}
\usetikzlibrary{pgfplots.groupplots}
\usetikzlibrary{
  pgfplots.colorbrewer,
}
\pgfplotsset{
  cycle list/.define={my marks}{
    every mark/.append style={solid,fill=\pgfkeysvalueof{/pgfplots/mark list fill}},mark=*\\
    every mark/.append style={solid,fill=\pgfkeysvalueof{/pgfplots/mark list fill}},mark=square*\\
    every mark/.append style={solid,fill=\pgfkeysvalueof{/pgfplots/mark list fill}},mark=triangle*\\
    every mark/.append style={solid,fill=\pgfkeysvalueof{/pgfplots/mark list fill}},mark=diamond*\\
  },
}

\DeclareMathSymbol{\mh}{\mathord}{operators}{`\-}
\definecolor{aa}{rgb}{0.19,0.55,0.91}
\definecolor{bb}{rgb}{0.9,0.17,0.31}
\definecolor{cc}{rgb}{0.514,0.325,0.831}
\definecolor{dd}{rgb}{0.12, 0.3, 0.17}
\definecolor{ee}{rgb}{1.0, 0.75, 0.0}
\definecolor{ff}{rgb}{0.01, 0.28, 1.0}
\definecolor{gg}{rgb}{0.65, 0.16, 0.16}

\AtBeginDocument{%
  \providecommand\BibTeX{{%
    Bib\TeX}}}

\setcopyright{acmlicensed}
\copyrightyear{2018}
\acmYear{2018}
\acmDOI{XXXXXXX.XXXXXXX}
\acmConference[CIKM 'XX]{Make sure to enter the correct
  conference title from your rights confirmation email}{June 03--05,
  2018}{Woodstock, NY}
\acmISBN{978-1-4503-XXXX-X/2018/06}

\begin{document}

\title[\textsf{TAHB}: A Comprehensive Benchmark for Text-Attributed Hypergraph Learning]{\textsf{TAHB}: A Comprehensive Benchmark \\for Text-Attributed Hypergraph Learning}

\author{David Yoon Suk Kang}
\email{dyskang@cbnu.ac.kr}
\affiliation{%
  \institution{Chungbuk National University}
  \country{South Korea}
}

\author{JungHyun Kim}
\email{rlawjdgus246@hanyang.ac.kr}
\affiliation{%
  \institution{Hanyang University}
  \country{South Korea}
}

\author{Juhyun Jeon}
\email{jjh1012@hanyang.ac.kr}
\affiliation{%
  \institution{Hanyang University}
  \country{South Korea}
}

\author{Sang-Wook Kim}
\authornote{Corresponding author.}
\email{wook@hanyang.ac.kr}
\affiliation{%
  \institution{Hanyang University}
  \country{South Korea} 
}

\renewcommand{\shortauthors}{Trovato et al.}

\begin{abstract}
\textit{Hypergraphs} effectively model higher-order groupwise relationships beyond pairwise interactions, while pretrained language models (PLMs) and large language models (LLMs) provide rich semantic understanding from textual attributes. However, research on combining language models with hypergraph learning remains limited due to the lack of public text-attributed hypergraph benchmarks.
To address this limitation, we present \textsf{TAHB} (Text-Attributed Hypergraph Benchmark), the first public benchmark integrating hypergraph structures and raw textual attributes. \textsf{TAHB} contains 10 real-world datasets from four domains—e-commerce, academia, movies, and politics networks—enabling systematic evaluation of text-aware hypergraph representation learning.
Experimental results show that \textsf{TAHB} preserves key structural properties of real-world hypergraphs and consistently reproduces performance tendencies observed in existing benchmarks. Furthermore, experiments under both LLM-as-Enhancer and LLM-as-Predictor settings demonstrate that LLM-enhanced textual semantics improve hypergraph learning performance, while structural and textual information jointly provide the best setting for LLM-based prediction. Our benchmark provides a foundation for future research at the intersection of hypergraph learning and language models.
\end{abstract}



\newcommand{\ie}{{\it i.e.}}
\newcommand{\eg}{{\it e.g.}}
\newcommand{\tahb}{\textsf{TAHB}}

\maketitle
\section{Introduction}~\label{s1}

\vspace{-3mm}
\noindent \textbf{Background and Motivation.} \textit{Graphs} have been widely adopted to represent relational and structural information across diverse domains~\cite{jang23:cikm, yoo23:www, new14:pnas, zh20:aaai, he21:icml, ya12:icdm}, and graph-based learning models have achieved remarkable success in numerous real-world applications. 

\begin{figure}[t]
\centering
\includegraphics[width=0.98\columnwidth]{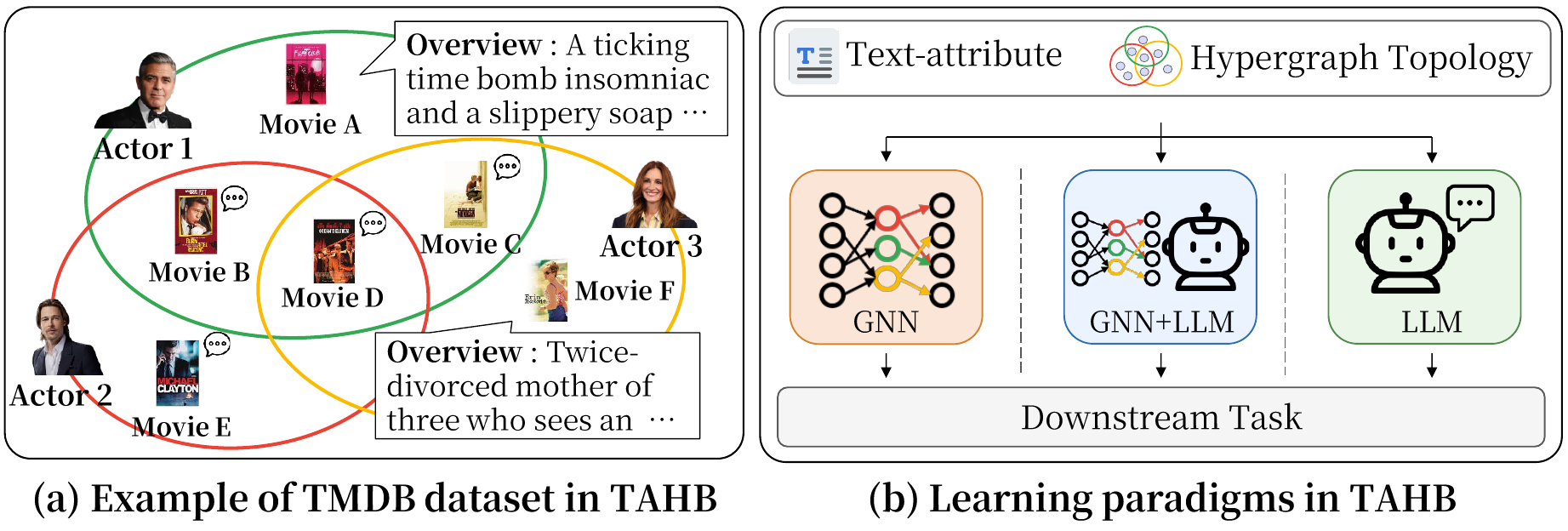}
\vspace{-4mm}
\caption{Overview of {\tahb}.}
\vspace{-7mm}
\label{Fig1}
\end{figure}

However, conventional graph learning research has primarily focused on topological structures, while recent studies suggest that structural information alone is often insufficient for capturing rich semantic relationships in real-world graphs~\cite{zhu20:neurips, xu2019:iclr, yang22:www}. 
As a result, increasing attention has been devoted to incorporating auxiliary semantic information such as node attributes~\cite{yang22:www}, edge signs~\cite{kan21:icdm}, textual semantics~\cite{jin24:tkde}, and multimodal signals~\cite{pen24:arxiv}.
In particular, whereas earlier studies represented textual attributes using relatively simple techniques such as one-hot encoding, TF--IDF, and Word2Vec~\cite{spa72:jd, mik13:arxiv}, recent advances in large language models (LLMs) have enabled highly expressive semantic representations~\cite{bro20:neurips, raf20:jmlr, tou23:arxiv}.  
This paradigm shift has stimulated active research on text-attributed graphs (TAGs)~\cite{jin23:acl, yan23:neurips, zha24:neurips}, where jointly leveraging graph structures and textual semantics consistently outperforms structure-only approaches.

Despite these advances, conventional graphs fundamentally rely on \textit{pairwise}, limiting their ability to model high-order interactions among multiple entities~\cite{zhou06:nips, do20:kdd, zha20:iclr}.  
\textit{Hypergraphs} address this limitation by allowing a single hyperedge to connect multiple nodes, thereby effectively preserving \textit{groupwise relationships}~\cite{zhou06:nips, do20:kdd}.  
Leveraging this expressive capability, hypergraph-based methods have achieved superior performance across various applications~\cite{han23:web, yu21:www, fen19:aaai, chi21:iclr, vel20:kdd}.

Nevertheless, the integration of LLMs into hypergraph learning remains severely limited~\cite{han23:web, yu21:www, fen19:aaai, chi21:iclr, vel20:kdd}, primarily due to the scarcity of real-world hypergraph datasets enriched with textual attributes.  
The absence of standardized text-attributed hypergraph benchmarks has therefore become a major bottleneck for systematically developing and evaluating LLM-driven hypergraph learning approaches.

\vspace{1mm}
\noindent \textbf{Proposed Benchmark.} Motivated by this challenge, we propose \textbf{{\tahb} (Text-Attributed Hypergraph Benchmark)}, the first publicly available benchmark that jointly incorporates textual attributes and hypergraph structures.
{\tahb} consists of 10 real-world hypergraph datasets (Figure 1-(a)) spanning four domains: e-commerce, academia, movies, and politics.  
Nodes represent entities such as papers, products, movies, and bills, while hyperedges capture naturally occurring groupwise relationships, including co-authorship, shared reviewers, common actors, and co-sponsorship relations.  
In addition, each node is associated with rich raw text information, such as paper abstracts, product descriptions and reviews, movie metadata, and bill contents.  
Consequently, {\tahb} provides a standardized evaluation framework for benchmarking both traditional hypergraph models and LLM-based approaches on key downstream tasks, including node classification and hyperedge prediction (Figure 1-(b)).

To validate the reliability of {\tahb}, we conduct extensive experiments from multiple perspectives.  
Our results show that {\tahb} preserves structural characteristics consistent with existing hypergraph benchmarks (having no text-attributes), while its text information exhibits distributions similar to those of established TAG datasets~\cite{yan23:neurips}.  
We further verify strong text--node semantic consistency, demonstrating that the associated text accurately describes its corresponding nodes.  
Finally, downstream task experiments reveal that performance trends on {\tahb} are highly consistent with those observed on existing hypergraph benchmarks (having no text-attributes), validating its reliability and practical utility as a standardized benchmark for text-attributed hypergraph learning.

\vspace{1mm}
\noindent \textbf{Potential of Integration with LLMs.} Furthermore, we investigate the integration of hypergraph representation learning and LLMs from two complementary perspectives~\cite{jin24:tkde}: \textit{LLM-as-Predictor}, where LLMs directly perform downstream tasks using hypergraph topology and textual semantics as input, and \textit{LLM-as-Enhancer}, where LLM-generated semantic augmentation is incorporated into existing HGNNs as enhanced input features.

Experimental results show that, in the \textit{LLM-as-Predictor} setting, jointly leveraging hypergraph topology and textual semantics consistently achieves the best performance across \textit{all} datasets, demonstrating the strong complementarity between the two modalities.  
In the \textit{LLM-as-Enhancer} setting, incorporating LLM-generated semantic augmentation consistently improves the performance of existing HGNN models over that of using only raw textual attributes.
These findings highlight the strong potential of integrating LLM-driven semantic information with hypergraph topology, and establish {\tahb} as a practical benchmark for future research on text-attributed hypergraph learning.

\vspace{1mm}
\noindent \textbf{Contributions.} The contributions of this paper are summarized as follows:
\vspace{-2mm}
\begin{itemize}[leftmargin=*]
\item \textbf{Novel Benchmark.} We introduce {\tahb}, the first publicized hypergraph benchmark jointly integrating textual attributes and hypergraph topology across 10 real-world datasets from four domains.
\item \textbf{Comprehensive Validation.} We verify that {\tahb} preserves real-world structural and semantic characteristics while reproducing consistent performance trends across downstream tasks.
\item \textbf{LLM--Hypergraph Integration.} We systematically investigate the integration of LLMs in hypergraph learning through \textit{LLM-as-Predictor} and \textit{LLM-as-Enhancer}, demonstrating the strong complementarity between LLM-driven semantics and hypergraph topology.
\end{itemize}

\vspace{1mm}
\noindent \textbf{Significance.} This work establishes a new research direction at the intersection of hypergraph learning and natural language processing by jointly modeling high-order structures and semantic information.  
By publicly releasing {\tahb} with standardized evaluation protocols, we provide a practical foundation for future research on LLM-driven text-attributed hypergraph learning.

\vspace{1mm}
\noindent \textbf{Relevance to the CIKM Research Track.} This work aligns with the CIKM Research Track topic of \textit{Evaluation}, including benchmarks, evaluation of generative and LLM-based systems, and reproducibility. In addition, prior CIKM research has continuously explored foundational datasets and benchmarking infrastructures~\cite{neo24:cikm, ni25:cikm}. In this work, we propose {\tahb}, a benchmark for evaluating hypergraph learning over text-attributed higher-order structures via language model integration.

\vspace{1mm}
\noindent \textbf{Reproducibility.} 
The experiments are conducted on the Linux server equipped with an AMD 7960X, 2TB NVMe SSD, 128GB DDR5 memory, and GeForce RTX A6000(48G).
For HRL methods, we use the source code provided by the authors of the original papers~\cite{fen19:aaai,yad19:neurips,don20:icmlw,hua21:ijcai,ci21:iclr,lee23:aaai,kim24:iclr}.
For parameters of each method, we use the best setting found via extensive grid search in the ranges suggested in its respective paper.
To ensure reproducibility, we make all source codes, datasets, and experimental pipelines for 10 real-world datasets publicly available at: https://anonymous.4open.science/r/TAHB-01F8
\section{Related Work}~\label{s2}

\vspace{-3mm}
\noindent \textbf{Hypergraph learning.}\label{2.1}
Existing HRL methods can be broadly divided into semi-supervised and self-supervised approaches.
Semi-supervised methods mainly extend Graph Neural Networks~\cite{kip17:iclr, vel17:arxiv} by transforming hypergraphs into graph-like structures, such as cliques~\cite{fen19:aaai,yad19:neurips} or bipartite graphs~\cite{don20:icmlw}, while UniGNN~\cite{hua21:ijcai} and AllSet~\cite{ci21:iclr} further generalize message passing and set-based hyperedge modeling.
In contrast, self-supervised HRL methods derive supervision directly from hypergraph topology~\cite{lee23:aaai, kim24:iclr}; for example, TriCL~\cite{lee23:aaai} employs tri-directional contrastive learning, whereas HypeBoy~\cite{kim24:iclr} adopts a partial hyperedge reconstruction strategy.

Despite their success, existing HRL methods have primarily been designed for structural features or shallow node embeddings, while the integration with \textit{Pre-trained Language Models} (PLMs) and \textit{Large Language Models} (LLMs) remains largely underexplored.
As a result, these methods are limited in their ability to capture the rich semantic information naturally embedded in real-world higher-order group interactions.

\begin{figure}[t]
\centering
\includegraphics[width=0.92\columnwidth]{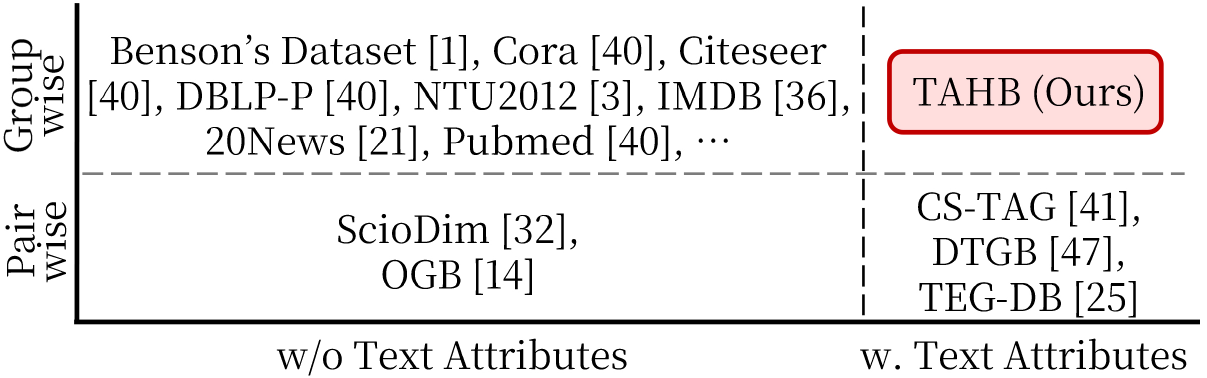}
\vspace{-3mm}
\caption{Taxonomy of existing graph and hypergraph benchmark datasets.}
\vspace{-6mm}
\label{Fig2}
\end{figure}

\begin{table*}[t]
\footnotesize
\centering

\caption{Comparison of {\tahb} and existing hypergraph benchmark datasets.}
\label{tab1}
\vspace{-4mm}
\def\arraystretch{1.3} 
\resizebox{\textwidth}{!}{\begin{tabular}{ l | c |r r r r c c c}
\toprule
 & \textbf{Datasets} & \textbf{Scale} &\textbf{\# of Nodes} & \textbf{\# of Hyperedges} & \textbf{\# of Classes} &
 \textbf{Domain} & \textbf{Feature Representation} & \textbf{Raw Text} \\
\midrule
\midrule
\multirow{12}{*}{\textbf{\shortstack{Existing}}} 
 & \textbf{Cora}~\cite{yad19:neurips} &Small& 1,434 & 1,579 & 7 & Academia & Bag-of-words & $\times$ \\
 & \textbf{Citeseer}~\cite{yad19:neurips} &small& 1,458 & 1,079 & 6 & Academia & Bag-of-words & $\times$ \\
 & \textbf{Cora-CA}~\cite{yad19:neurips} & Small&2,388 & 1,072 & 7 & Academia & Bag-of-words & $\times$ \\
 & \textbf{DBLP-A}~\cite{wan19:www} & Small&2,591 & 2,690 & 4 & Academia & Bag-of-words & $\times$ \\
 & \textbf{Pubmed}~\cite{yad19:neurips} & Small&3,840 & 7,963 & 3 & Academia & Bag-of-words & $\times$ \\
 & \textbf{AMiner}~\cite{zha19:kdd} & Large&20,201 & 8,052 & 12 & Academia & Bag-of-words & $\times$ \\
 & \textbf{DBLP-P}~\cite{yad19:neurips} & Large &41,302 & 22,263 & 6 & Academia & Bag-of-words & $\times$ \\
 & \textbf{NTU2012}~\cite{che03:cg} &Small& 2,012 & 2,012 & 67 & Computer Graphics & GVCNN & $\times$ \\
 & \textbf{ModelNet40}~\cite{wu15:cvpr} & Medium&12,311 & 12,311 & 40 & Computer Graphics & GVCNN & $\times$ \\
 & \textbf{20News} & Medium&16,242 & 100 & 4 & News & TF-IDF & $\times$ \\
 & \textbf{IMDB}~\cite{wan19:www} & Small&3,939 & 2,015 & 3 & Movies & Bag-of-words & $\times$ \\
 & \textbf{House}~\cite{ci21:iclr} & Small&1,290 & 341 & 2 & Politics & One-hot encoding & $\times$ \\

\hline
\multirow{10}{*}{\textbf{\shortstack{{\tahb}}}} 
 & \textbf{Amazon Pet Supplies} &Medium& 10,466 & 28,441 & 5 & E-commerce & PLMs & $\bigcirc$ \\
 & \textbf{Amazon Arts \& Crafts}  &Medium& 14,728 & 50,908 & 7 & E-commerce & PLMs & $\bigcirc$ \\
 & \textbf{Amazon Grocery} &Medium & 19,272 & 56,667 & 5 & E-commerce & PLMs & $\bigcirc$ \\
 & \textbf{Amazon Electronics} &Large& 23,149 & 40,626 & 3 & E-commerce & PLMs & $\bigcirc$ \\
 & \textbf{arXiv Math} & Small& 4,118 & 2,342 & 3 & Academia & PLMs & $\bigcirc$ \\
 & \textbf{arXiv Physics} &Large& 38,446 & 49,907 & 3 & Academia & PLMs & $\bigcirc$ \\
 & \textbf{arXiv CS} &Large& 60,771 & 37,714 & 4 & Academia & PLMs & $\bigcirc$ \\
 & \textbf{TMDB} &Large & 45,319 & 21,404 & 19 & Movies & PLMs & $\bigcirc$ \\
 & \textbf{Senate} & Small& 9,512 & 411 & 2 & Politics & PLMs & $\bigcirc$ \\
 & \textbf{House} &Medium& 12,281 & 1,930 & 2 & Politics & PLMs & $\bigcirc$ \\
 
\bottomrule
\end{tabular}
}
\vspace{-3mm}
\end{table*}

\vspace{1mm}
\noindent\textbf{Benchmark Datasets in Graph and Hypergraph Learning.}
Recent advances in graph learning have increasingly emphasized the importance of incorporating semantic information beyond pure topology. 
Accordingly, existing benchmark datasets can be broadly categorized from two perspectives: (1) whether they provide textual attributes and (2) whether they model primitive pairwise relations or compressive higher-order relations. 
Traditional graph benchmarks such as SocioDim~\cite{tang09:kdd} and OGB~\cite{hu20:neurips} mainly focus on primitive pairwise structures without rich textual semantics. 
More-recent text-attributed benchmarks, including CS-TAG~\cite{yan23:neurips}, DTGB~\cite{zha24:neurips}, and TEG-DB~\cite{he24:neurips}, provide rich textual information but still rely on standard pairwise graph formulations. 
In contrast, existing hypergraph benchmarks, such as Benson’s datasets~\cite{ben18:pnas} and widely used benchmarks for node classification and hyperedge prediction (\eg, Cora~\cite{yad19:neurips}, Citeseer~\cite{yad19:neurips}, Cora-CA~\cite{yad19:neurips}, DBLP-A~\cite{wan19:www}, Pubmed~\cite{yad19:neurips}, AMiner~\cite{zha19:kdd}, DBLP-P~\cite{yad19:neurips}, NTU2012~\cite{che03:cg}, ModelNet40~\cite{wu15:cvpr}, 20News, IMDB~\cite{wan19:www}, and House~\cite{ci21:iclr}), effectively capture higher-order group relations through hypergraph structures, yet generally lack raw textual attributes associated with nodes.

\noindent\textbf{Benchmark Gap and Our Positioning.}
As illustrated in Figure~\ref{Fig2}, existing datasets only partially satisfy the requirements for modern text-aware hypergraph learning. 
While recent graph benchmarks provide rich textual semantics, they mainly rely on primitive pairwise structures. 
Conversely, existing hypergraph benchmarks capture compressive higher-order relations but generally lack raw textual attributes. 
Consequently, there remains no public benchmark that jointly supports both higher-order relational modeling and raw text-aware semantic learning. 
To address this gap, we propose {\tahb}, the first benchmark specifically designed for text-attributed hypergraph learning.


\vspace{-5mm}
\section{{\tahb}: The Proposed Benchmark}~\label{s3}
In this section, we present the construction of {\tahb} and its key characteristics.
First, Section 3.1 provides an overview of the overall architecture and composition of {\tahb}.
Next, Section 3.2 describes the data collection and preprocessing procedures, including the constituent elements of each dataset such as nodes and hyperedges.
Finally, Section 3.3 validates {\tahb} as a benchmark for text-attributed hypergraph learning through analyses of (1) \textit{structural characteristics}, (2) \textit{textual attribute distributions}, and (3) \textit{performance tendencies} of various hypergraph learning methods compared with those of existing hypergraph datasets.

\subsection{Overview}
As summarized in Table~\ref{tab1}, {\tahb} comprises 10 real-world datasets spanning four diverse domains: E-commerce, Academia, Movies, and Politics.
Each domain is selected because textual attributes and higher-order relational structures naturally coexist, enabling the evaluation of model generalizability across diverse real-world scenarios.

Based on the node scale distributions of existing hypergraph benchmark datasets~\cite{yad19:neurips,wan19:www,zha19:kdd,che03:cg,wu15:cvpr,ci21:iclr}, {\tahb} includes three dataset scales: small (1,000–10,000 nodes), medium (10,000–20,000 nodes), and large (more than 20,000 nodes), thereby supporting scalability evaluations across varying graph sizes.

Existing hypergraph benchmarks often rely on numerical metadata or preprocessed shallow features, which may fail to preserve the semantics inherent in raw text. 
To address this limitation, {\tahb} preserves the original raw text associated with all nodes in the hypergraph. 
This design choice enables researchers to flexibly integrate pretrained language models (PLMs) or large language models (LLMs) for advanced textual representation learning and domain-specific feature extraction.

Furthermore, to maximize reproducibility, we standardize the storage formats of hyperedges and textual attributes and provide a modular experimental pipeline that directly supports node classification and hyperedge prediction tasks. 
Such a standardized environment facilitates fair comparisons across models and establishes a systematic benchmark foundation for future research on text-attributed hypergraph learning.

\subsection{Dataset Construction}\label{3.2}
\noindent \textbf{Dataset Collection.}
To construct TAHB, we collect text-attributed hypergraph datasets from four real-world domains: E-commerce, Academia, Movies, and Politics.  
Each dataset consists of nodes associated with raw textual attributes and hyperedges representing naturally occurring higher-order group relationships.  
The detailed descriptions of each dataset are as follows:
\begin{itemize}[leftmargin=*]
\item \textbf{E-commerce.}
Collected from the \textit{Amazon Review dataset}~\cite{hou24:arxiv} and further processed into hypergraph structures using the \textit{Pet Supplies}, \textit{Arts \& Crafts}, \textit{Grocery}, and \textit{Electronics} categories.
Each node represents a product, where the node text consists of the product title and description, and the node label corresponds to the product sub-category.
A hyperedge indicates as the set of products purchased and reviewed by the same user.

\item \textbf{Academia.}
Collected from \textit{arXiv} using papers from the \textit{Computer Science}, \textit{Physics}, and \textit{Mathematics} categories.\footnote{https://www.arxiv.org}
Each node represents a paper, where the node text consists of the paper title and abstract, and the node label corresponds to the paper sub-field.
A hyperedge indicates as the set of papers authored by the same researcher.

\item \textbf{Movies.}
Collected from movies in \textit{TMDB}.\footnote{https://www.themoviedb.org}
Each node represents a movie, where the node text consists of the movie title and plot description, and the node label corresponds to the movie genre.
A hyperedge indicates as the set of movies featuring the same actor.

\item \textbf{Politics.}
Collected from bills in \textit{GovTrack} proposed in the U.S. Senate and House of Representatives.\footnote{https://www.govtrack.us}
Each node represents a bill, where the node text consists of the bill title and summary, and the node label indicates whether the bill was passed or rejected.
A hyperedge is defined as the set of bills proposed by the same legislator.

\end{itemize}

\begin{table*}[t]
\footnotesize

\caption{Structural characteristics of {\tahb} datasets.}
\label{tab6}
\vspace{-4mm}

\setlength{\tabcolsep}{4pt}
\renewcommand{\arraystretch}{1.15}

\resizebox{\textwidth}{!}{
\begin{tabular}{lcccccccccc}
\toprule

& \makecell{\textbf{Amazon} \\ \textbf{Pet Supplies}}
& \makecell{\textbf{Amazon} \\ \textbf{Arts \& Crafts}}
& \makecell{\textbf{Amazon} \\ \textbf{Grocery}}
& \makecell{\textbf{Amazon} \\ \textbf{Electronics}}
& \makecell{\textbf{arXiv} \\ \textbf{Math}}
& \makecell{\textbf{arXiv} \\ \textbf{Physics}}
& \makecell{\textbf{arXiv} \\ \textbf{CS}}
& \textbf{TMDB}
& \textbf{Senate}
& \textbf{House} \\

\midrule

\textbf{Connected Component Ratio}
& 1.00 & 1.00 & 1.00 & 0.99 & 0.70 & 0.97 & 0.97 & 0.97 & 1.00 & 1.00 \\

\textbf{Effective Diameter}
& 3.16 & 2.74 & 2.84 & 3.46 & 13.74 & 5.42 & 5.49 & 4.95 & 1.96 & 2.16 \\

\midrule

\textbf{Clustering Coefficient (Random)}
& 0.16$\pm$4e-4
& 0.12$\pm$2e-4
& 0.12$\pm$1e-4
& 0.24$\pm$8e-4
& 0.52$\pm$4e-3
& 0.31$\pm$4e-4
& 0.48$\pm$6e-4
& 0.58$\pm$1e-3
& 0.41$\pm$5e-4
& 0.30$\pm$3e-4 \\

\textbf{Clustering Coefficient ({\tahb})}
& 0.46 & 0.28 & 0.35 & 0.52 & 0.71 & 0.53 & 0.64 & 0.59 & 0.57 & 0.45 \\

\bottomrule
\end{tabular}
}
\vspace{-4mm}
\end{table*}

\vspace{1mm}
\noindent \textbf{Data Preprocessing.}
To ensure the integrity and structural validity of {\tahb}, the collected raw data are refined through the following three-stage preprocessing pipeline.
First, we remove nodes whose textual attributes are missing or contain noisy symbols and non-English sentences that may hinder contextual learning.
Second, to rigorously preserve the higher-order relational properties of hypergraphs, we remove singleton hyperedges and isolated nodes that do not belong to any hyperedge.
Finally, all preprocessed datasets are stored in a unified format consisting of node textual attributes, hypergraph structural information, and ground-truth labels.

\subsection{Dataset Validation}\label{3.3}
\noindent \textbf{Validation Questions.} In this section, we validate whether the constructed {\tahb} exhibits the characteristics of real-world hypergraph data and can serve as a reliable benchmark for hypergraph mining research.
The validation is designed to answer the following validation questions (VQs):

\begin{itemize}[leftmargin=*]
\item \textbf{(VQ1)} Does {\tahb} exhibit structural characteristics similar to those of existing real-world hypergraph datasets?
\item \textbf{(VQ2)} Do the textual attributes in {\tahb} exhibit realistic distributions and informative semantics?
\item \textbf{(VQ3)} Are the performance tendencies of HRL methods consistently reproduced on {\tahb}?
\end{itemize}

\noindent \textbf{VQ1. Structural Characteristics.}
Real-world hypergraphs are known to exhibit the following six structural characteristics~\cite{do20:kdd, ben18:pnas}: \textbf{(S1)} high giant-connected-component ratio, \textbf{(S2)} small effective diameter, \textbf{(S3)} high clustering coefficient, \textbf{(S4)} heavy-tailed degree distribution, \textbf{(S5)} heavy-tailed singular value distribution, and \textbf{(S6)} heavy-tailed hyperedge size distribution.

Following~\cite{do20:kdd}, we first transform each hypergraph into a standard graph via \textit{clique expansion}~\cite{sun08:kdd} and then analyze the five structural characteristics on the transformed graph.
Here, the effective diameter (S2) is defined as the minimum distance $d$ such that approximately 90\% of all connected node pairs can be reached within a path length of at most $d$.
For (S3), to verify that the observed clustering coefficients are not caused by random structural effects, we additionally compare each real-world hypergraph with its corresponding random hypergraph, following the protocol of~\cite{do20:kdd}.
Specifically, for each original hypergraph, we construct a random hypergraph of the same size (\ie, same numbers of nodes and hyperedges) by randomly selecting the nodes contained in each hyperedge.

\begin{figure*}[t]
\centering
\includegraphics[width=0.95\textwidth]{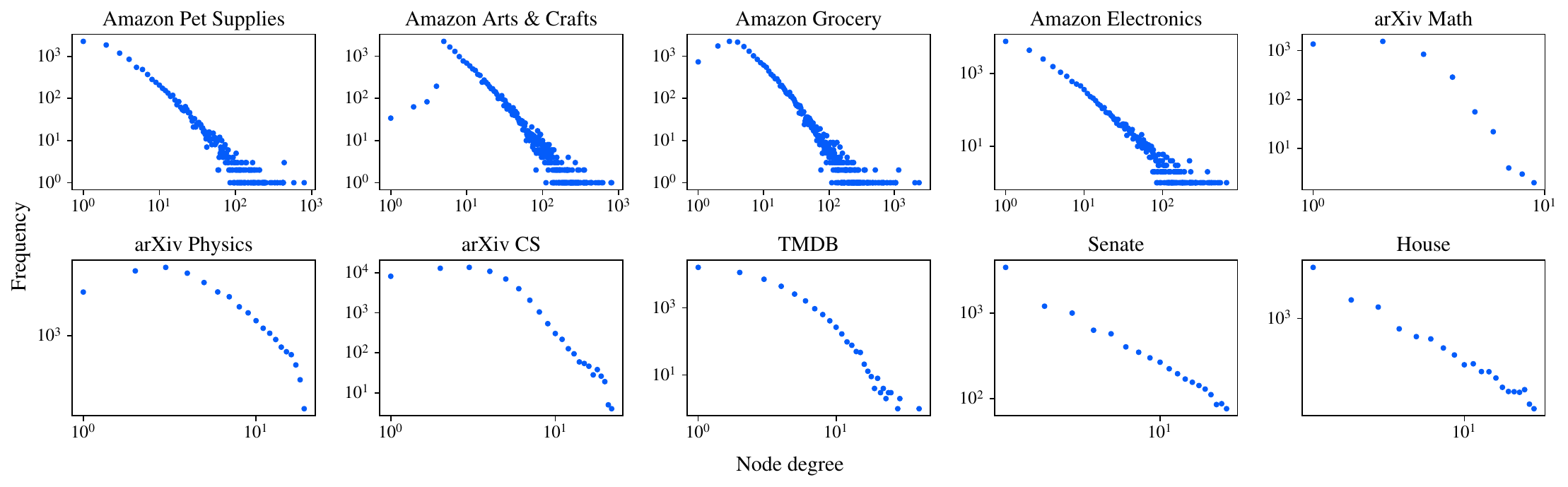}
\vspace{-5mm}
\caption{Distribution of node degrees across the {\tahb} datasets.}
\vspace{-1mm}
\label{Fig3}
\end{figure*}

\begin{figure*}[t]
\centering
\includegraphics[width=0.95\textwidth]{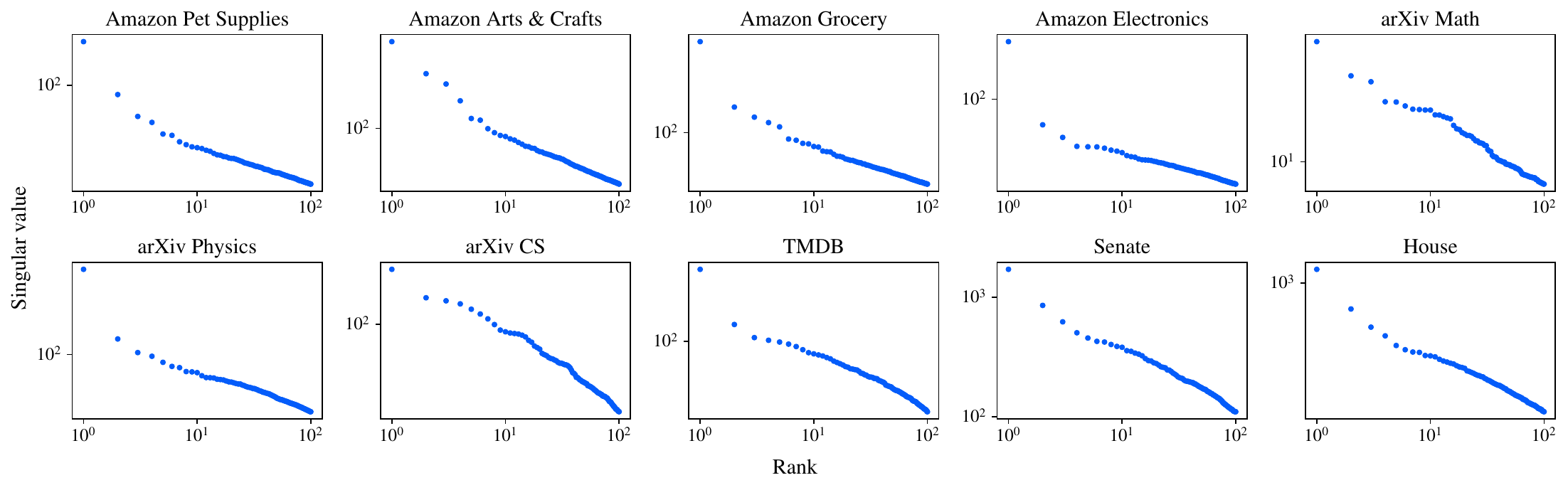}
\vspace{-5mm}
\caption{Distribution of singular-values across the {\tahb} datasets.}
\vspace{-1mm}
\label{Fig4}
\end{figure*}

\begin{figure*}[t]
\centering
\includegraphics[width=0.95\textwidth]{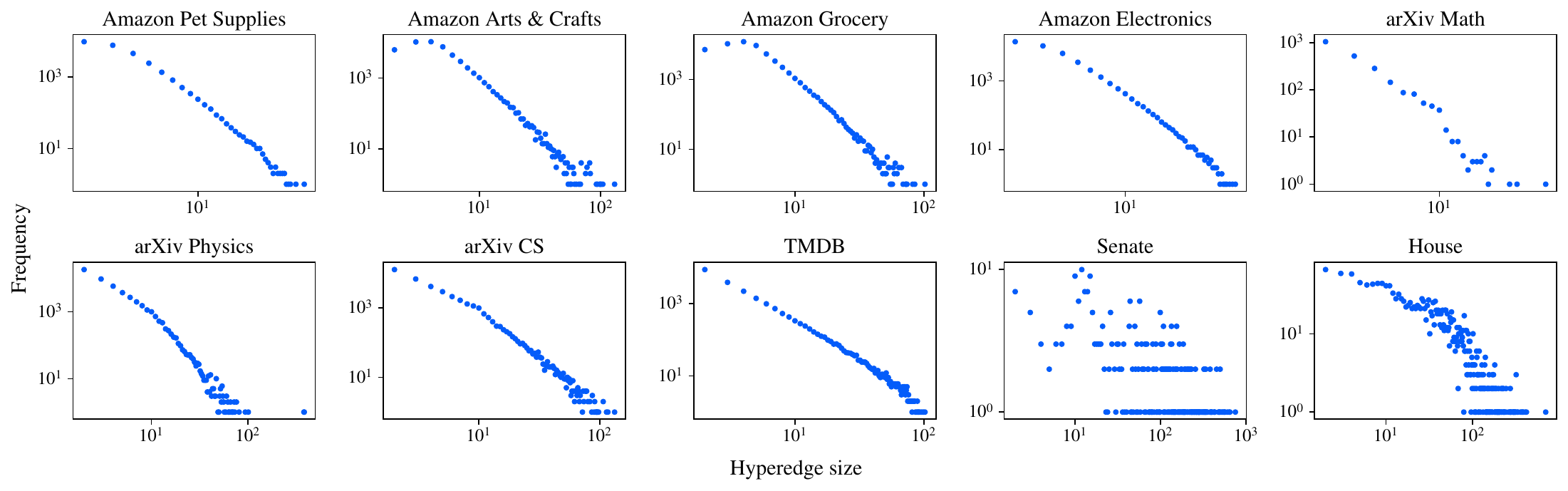}
\vspace{-5mm}
\caption{Distribution of hyperedge sizes across the {\tahb} datasets.}
\vspace{-3mm}
\label{Fig8}
\end{figure*}

Table~2 reports the giant connected component ratio \textbf{(S1)}, effective diameter \textbf{(S2)}, and clustering coefficient \textbf{(S3)} for each dataset.
For \textbf{(S1)}, consistent with the observations in~\cite{do20:kdd}, we find that the vast majority of nodes in the {\tahb} datasets belong to a giant connected component.
For \textbf{(S2)}, again following the trends reported in~\cite{do20:kdd}, the effective diameters of most datasets range between 2 and approximately 10, indicating a short reachability between nodes.
For \textbf{(S3)}, the real-world hypergraphs in {\tahb} consistently exhibit higher clustering coefficients than their corresponding random hypergraphs, which aligns well with the findings in~\cite{do20:kdd}.
Figure~\ref{Fig3}, Figure~\ref{Fig4}, and Figure~\ref{Fig8} present the degree distributions \textbf{(S4)}, singular value distributions \textbf{(S5)}, and hyperedge size distributions \textbf{(S6)}, respectively.
Similar to existing real-world hypergraph datasets, the degree, singular value, and hyperedge size distributions in {\tahb} consistently exhibit clear heavy-tailed patterns.

These observations collectively demonstrate that {\tahb} preserves the fundamental structural characteristics of real-world hypergraphs.
This suggests that {\tahb} is not merely a hypergraph dataset augmented with textual attributes, but a realistic benchmark that integrates textual information while maintaining real-world hypergraph characteristics.

\vspace{1mm}
\noindent \textbf{VQ2. Text Attribute Distribution and Quality.}
In this section, we validate the textual attributes collected in {\tahb}.
In real-world text-attributed graphs, node textual attributes are generally known to follow right-skewed distributions~\cite{zha24:neurips}.
We investigate whether the same tendency is also observed in {\tahb}.\footnote{Since no benchmark currently exists for text-attributed hypergraphs, we compare the textual characteristics of {\tahb} with those observed in text-attributed graphs (TAGs). Hypergraphs are generally regarded as generalized forms of graphs and are known to exhibit structural tendencies similar to those of standard graphs~\cite{do20:kdd}.}

\begin{figure*}[t]
\centering
\includegraphics[width=0.95\textwidth]{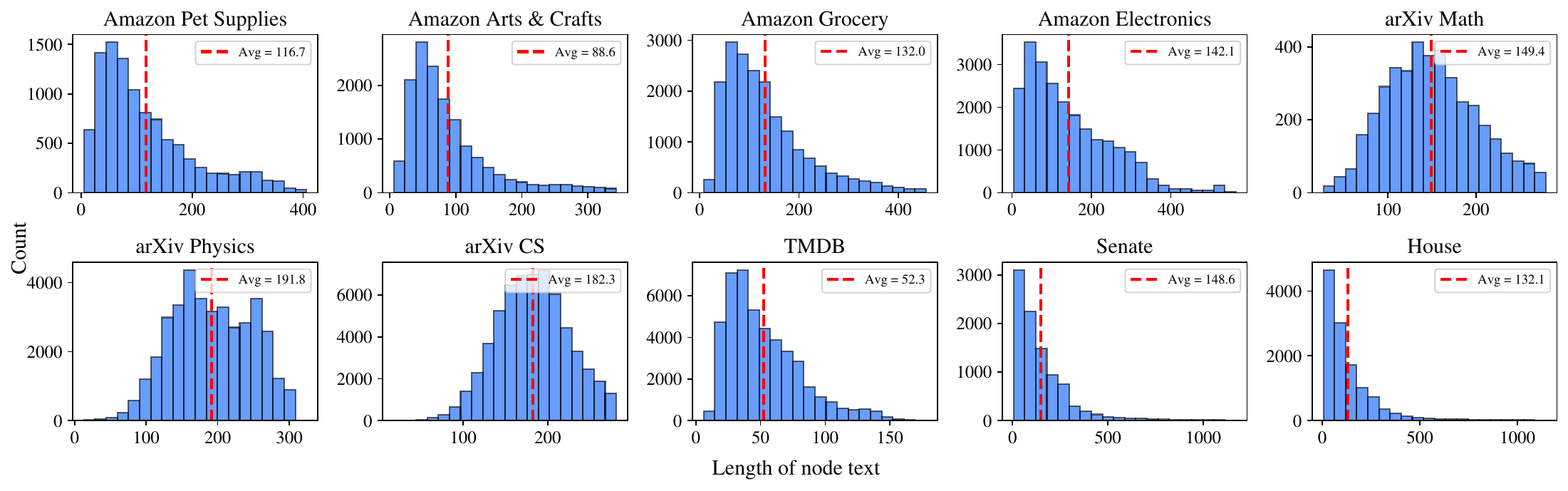}
\vspace{-5mm}
\caption{Distribution of node text lengths across the {\tahb} datasets.}
\vspace{-3mm}
\label{Fig5}
\end{figure*}

Figure~\ref{Fig5} presents the distributions of node text lengths for each dataset.
As shown in the figure, {\tahb} exhibits the right-skewed distributions commonly observed in real-world TAG datasets.
These results indicate that the textual attributes collected in {\tahb} well preserve the general characteristics of real-world text-attributed data.

Next, we investigate whether the collected textual attributes contain informative semantics that appropriately characterize individual nodes. To this end, we conduct node classification as a downstream task and compare the performance of randomly initialized features with that of text-based features encoded by a pretrained language model (PLM). Specifically, we use BERT-Tiny as the PLM encoder and HGNN~\cite{fen19:aaai} as the backbone HRL model. Intuitively, if the collected textual attributes effectively describe the nodes, the text-based features should consistently outperform the random features; otherwise, their performance would be comparable or even worse.

\begin{figure*}[t]
\centering
\includegraphics[width=0.95\textwidth]{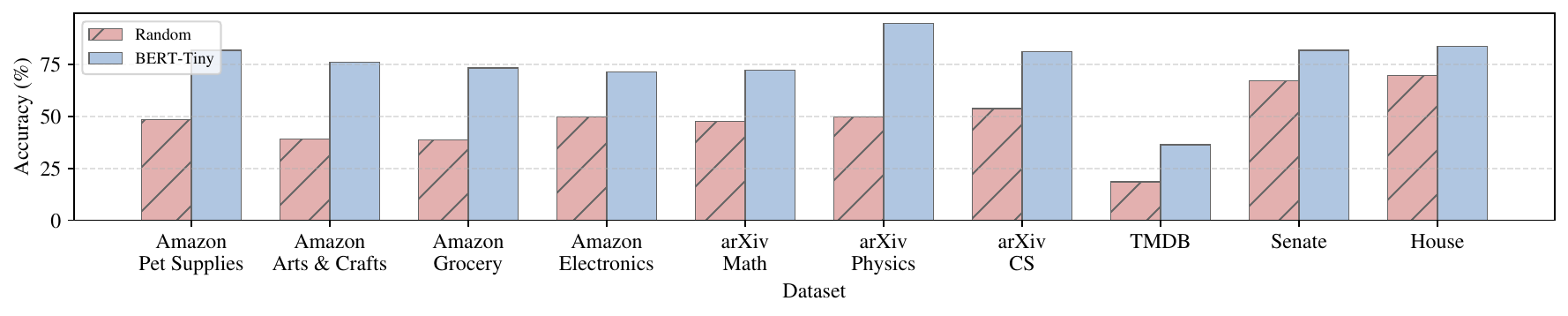}
\vspace{-5mm}
\caption{Comparison of node classification accuracies between random and text-attributed features on the {\tahb} datasets.}
\vspace{-3mm}
\label{Fig88}
\end{figure*}

Figure~\ref{Fig88} shows the corresponding results. 
Compared with random features, the text-based features consistently achieve substantially higher performance on the node classification task across all datasets.\footnote{We observe similar trends across other HRL methods as well.} 
These results demonstrate that the textual attributes in {\tahb} contain informative semantics that effectively characterize individual nodes.

\vspace{1mm}
\noindent \textbf{VQ3. Performance Tendency Reproducibility.}
Finally, we investigate whether the performance trends of various hypergraph representation learning (HRL) methods observed on existing hypergraph benchmarks are consistently reproduced on {\tahb}. 
To this end, we conduct two downstream tasks, node classification and hyperedge prediction, following the experimental protocols of prior studies~\cite{kim24:iclr}. 
We evaluate seven state-of-the-art HRL methods: HGNN~\cite{fen19:aaai}, HyperGCN~\cite{yad19:neurips}, HNHN~\cite{don20:icmlw}, AllSet~\cite{ci21:iclr}, UniGCNII~\cite{hua21:ijcai}, TriCL~\cite{lee23:aaai}, and HypeBoy~\cite{kim24:iclr}.

\begin{table*}[t]
\centering
\footnotesize
\renewcommand{\arraystretch}{1.15}
\setlength{\tabcolsep}{5pt}
\caption{Node classification accuracy of HRL methods under different text encoders on {\tahb}.}
\label{ttab2}
\vspace{-4mm}
\begin{tabular*}{\textwidth}{@{\extracolsep{\fill}}ll||ccccccc}
\toprule
\textbf{Dataset} & \textbf{Embedding}
& \textbf{HGNN} & \textbf{HyperGCN} & \textbf{HNHN} & \textbf{Allset} & \textbf{UniGCNII} & \textbf{TriCL} & \textbf{HypeBoy} \\
\midrule\midrule

\multirow{4}{*}{\textbf{Amazon Pet Supplies}}
& \textbf{Skip-Gram}
&61.37$\pm$0.58 (5)&61.02$\pm$0.27 (6)&62.92$\pm$0.90 (3)&61.02$\pm$0.27 (7)&62.67$\pm$1.42 (4)&\underline{63.21$\pm$1.11} (2)&\textbf{64.33$\pm$1.54} (1)\\
& \textbf{BERT-Tiny}
&81.97$\pm$0.43 (5)&79.35$\pm$0.69 (7)&80.14$\pm$0.54 (6)&82.89$\pm$0.75 (3)&82.86$\pm$1.22 (4)&\textbf{85.97$\pm$1.36} (1)&\underline{85.83$\pm$3.10} (2)\\
& \textbf{BERT-Base}
&82.86$\pm$1.19 (5)&80.91$\pm$5.23 (7)&82.25$\pm$0.42 (6)&84.61$\pm$0.82 (4)&86.77$\pm$0.42 (3)&\underline{88.89$\pm$1.26} (2)&\textbf{89.13$\pm$1.12} (1)\\
& \textbf{DeBERTa-Large}
&70.01$\pm$1.79 (5)&68.53$\pm$1.95 (7)&69.03$\pm$0.88 (6)&75.42$\pm$1.42 (3)&75.13$\pm$0.81 (4)&\textbf{79.88$\pm$0.06} (1)&\underline{78.86$\pm$0.08} (2)\\
\midrule

\multirow{4}{*}{\textbf{Amazon Arts \& Crafts}}
& \textbf{Skip-Gram}
&34.12$\pm$8.15 (6)&32.28$\pm$3.67 (7)&34.41$\pm$1.78 (5)&37.79$\pm$6.80 (4)&41.03$\pm$1.87 (3)&\textbf{46.37$\pm$3.37} (1)&\underline{45.79$\pm$1.11} (2)\\
& \textbf{BERT-Tiny}
&75.78$\pm$0.93 (5)&70.18$\pm$2.66 (7)&74.33$\pm$0.78 (6)&76.23$\pm$0.40 (4)&77.26$\pm$0.38 (3)&\underline{82.57$\pm$1.62} (2)&\textbf{82.99$\pm$3.11} (1)\\
& \textbf{BERT-Base}
&77.56$\pm$1.37 (5)&72.17$\pm$2.09 (7)&77.50$\pm$0.31 (6)&77.98$\pm$0.63 (4)&79.74$\pm$0.44 (3)&\underline{83.23$\pm$0.41} (2)&\textbf{83.82$\pm$3.44} (1)\\
& \textbf{DeBERTa-Large}
&68.92$\pm$1.74 (5)&59.78$\pm$1.12 (7)&68.76$\pm$0.71 (6)&70.82$\pm$0.73 (3)&70.57$\pm$1.11 (4)&\textbf{73.59$\pm$2.64} (1)&\underline{72.75$\pm$2.92} (2)\\
\midrule

\multirow{4}{*}{\textbf{Amazon Grocery}}
& \textbf{Skip-Gram}
&33.90$\pm$0.58 (5)&33.35$\pm$0.16 (6)&35.96$\pm$1.23 (4)&33.27$\pm$0.16 (7)&39.25$\pm$2.33 (3)&\underline{40.25$\pm$0.05} (2)&\textbf{40.76$\pm$0.49} (1)\\
& \textbf{BERT-Tiny}
&68.23$\pm$3.66 (6)&60.45$\pm$2.46 (7)&69.36$\pm$0.39 (4)&68.31$\pm$2.46 (5)&71.59$\pm$0.75 (3)&\underline{72.23$\pm$1.14} (2)&\textbf{73.54$\pm$1.04} (1)\\
& \textbf{BERT-Base}
&71.71$\pm$2.74 (6)&58.67$\pm$5.06 (7)&72.59$\pm$0.20 (5)&73.44$\pm$0.86 (3)&72.99$\pm$0.34 (4)&\textbf{74.82$\pm$1.65} (1)&\underline{74.28$\pm$0.87} (2)\\
& \textbf{DeBERTa-Large}
&56.89$\pm$2.19 (6)&43.33$\pm$6.12 (7)&57.70$\pm$1.46 (5)&58.99$\pm$1.85 (4)&59.87$\pm$1.56 (3)&\underline{63.47$\pm$0.26} (2)&\textbf{65.49$\pm$1.08} (1)\\
\midrule

\multirow{4}{*}{\textbf{Amazon Electronics}}
& \textbf{Skip-Gram}
&53.47$\pm$0.24 (6)&53.48$\pm$0.13 (5)&54.67$\pm$0.52 (3)&53.42$\pm$0.13 (7)&53.89$\pm$0.52 (4)&\underline{57.42$\pm$0.14} (2)&\textbf{58.32$\pm$0.32} (1)\\
& \textbf{BERT-Tiny}
&71.47$\pm$0.91 (5)&68.96$\pm$1.11 (7)&72.06$\pm$0.09 (4)&70.81$\pm$0.75 (6)&77.35$\pm$0.48 (3)&\underline{80.42$\pm$1.02} (2)&\textbf{82.95$\pm$0.91} (1)\\
& \textbf{BERT-Base}
&73.26$\pm$0.34 (6)&73.42$\pm$6.02 (5)&74.74$\pm$0.35 (4)&72.66$\pm$0.82 (7)&76.59$\pm$0.42 (3)&\underline{81.93$\pm$1.32} (2)&\textbf{82.51$\pm$0.46} (1)\\
& \textbf{DeBERTa-Large}
&63.94$\pm$0.79 (6)&57.06$\pm$1.24 (7)&65.49$\pm$0.77 (4)&64.28$\pm$1.75 (5)&67.41$\pm$0.74 (3)&\underline{68.41$\pm$0.04} (2)&\textbf{68.75$\pm$0.42} (1)\\
\midrule

\multirow{4}{*}{\textbf{arXiv Math}}
& \textbf{Skip-Gram}
&39.52$\pm$2.30 (7)&40.15$\pm$2.97 (5)&39.96$\pm$2.16 (6)&41.61$\pm$3.06 (3)&41.42$\pm$2.59 (4)&\textbf{43.80$\pm$0.95} (1)&\underline{43.59$\pm$3.06} (2)\\
& \textbf{BERT-Tiny}
&72.28$\pm$2.56 (5)&70.71$\pm$2.77 (7)&72.08$\pm$2.87 (6)&72.32$\pm$2.26 (4)&72.63$\pm$2.39 (3)&\textbf{76.48$\pm$2.40} (1)&\underline{75.49$\pm$0.90} (2)\\
& \textbf{BERT-Base}
&65.72$\pm$4.92 (5)&64.87$\pm$6.31 (7)&65.54$\pm$2.57 (6)&66.38$\pm$3.95 (4)&66.79$\pm$3.31 (3)&\textbf{70.66$\pm$2.52} (1)&\underline{70.09$\pm$3.52} (2)\\
& \textbf{DeBERTa-Large}
&65.97$\pm$1.86 (5)&62.10$\pm$2.37 (7)&64.84$\pm$0.46 (6)&66.05$\pm$1.20 (4)&68.21$\pm$0.38 (3)&\underline{68.41$\pm$0.04} (2)&\textbf{70.75$\pm$0.42} (1)\\
\midrule

\multirow{4}{*}{\textbf{arXiv Physics}}
& \textbf{Skip-Gram}
&\underline{62.62$\pm$0.86} (2)&61.75$\pm$0.77 (5)&60.52$\pm$2.53 (6)&61.96$\pm$0.64 (4) &62.23$\pm$2.27 (3)&\textbf{64.98$\pm$0.33} (1)&OOM\\
& \textbf{BERT-Tiny}
&95.06$\pm$0.16 (4)&93.62$\pm$2.87 (6)&94.67$\pm$0.20 (5)&95.25$\pm$0.14 (3)&\underline{95.30$\pm$0.19} (2)&\textbf{96.13$\pm$0.29} (1)&OOM\\
& \textbf{BERT-Base}
&94.28$\pm$0.42 (4)&92.14$\pm$3.22 (6)&93.88$\pm$0.69 (5)&94.36$\pm$0.55 (3)&\underline{94.70$\pm$0.36} (2)&\textbf{95.90$\pm$0.35} (1)&OOM\\
& \textbf{DeBERTa-Large}
&\underline{95.25$\pm$1.56} (2)&91.88$\pm$0.03 (6)&94.84$\pm$0.24 (5)&95.15$\pm$0.14 (4)&\textbf{95.38$\pm$0.08} (1) &95.20$\pm$0.79 (3) &OOM\\
\midrule

\multirow{4}{*}{\textbf{arXiv CS}}
& \textbf{Skip-Gram}
&40.34$\pm$0.10 (3)&39.20$\pm$0.12 (6)&40.31$\pm$0.05 (4)&40.30$\pm$0.03 (5)&\underline{40.45$\pm$0.14} (2)&\textbf{56.69$\pm$1.85} (1)&OOM\\
& \textbf{BERT-Tiny}
&81.19$\pm$0.22 (3)&40.53$\pm$0.13 (6)&80.37$\pm$0.13 (5)&81.15$\pm$0.16 (4)&\underline{84.38$\pm$0.26} (2)&\textbf{86.17$\pm$0.14} (1)&OOM\\
& \textbf{BERT-Base}
&82.36$\pm$2.64 (4)&41.20$\pm$0.09 (6)&82.07$\pm$0.23 (5)&82.74$\pm$0.22 (3)&\underline{87.55$\pm$0.08} (2)&\textbf{90.25$\pm$0.79} (1)&OOM\\
& \textbf{DeBERTa-Large}
&75.13$\pm$1.87 (3)&39.80$\pm$0.22 (6)&73.05$\pm$1.29 (5)&74.74$\pm$1.28 (4)&\underline{76.82$\pm$0.46} (2)&\textbf{80.46$\pm$0.40} (1)&OOM\\
\midrule

\multirow{4}{*}{\textbf{TMDB}}
& \textbf{Skip-Gram}
&23.18$\pm$1.26 (7)&24.04$\pm$0.07 (5)&24.05$\pm$0.10 (4)&24.03$\pm$0.07 (6)&24.09$\pm$0.08 (3)&\textbf{27.01$\pm$0.29} (1)&\underline{26.67$\pm$0.92} (2)\\
& \textbf{BERT-Tiny}
&35.35$\pm$2.16 (6)&33.67$\pm$0.57 (7)&35.37$\pm$0.13 (5)&36.43$\pm$0.48 (4)&42.83$\pm$0.19 (3)&\textbf{45.17$\pm$1.16} (1)&\underline{44.19$\pm$1.81} (2)\\
& \textbf{BERT-Base}
&37.20$\pm$0.78 (5)&36.87$\pm$1.31 (7)&37.15$\pm$0.31 (6)&37.31$\pm$0.61 (4)&47.12$\pm$0.28 (3)&\underline{49.05$\pm$1.08} (2)&\textbf{50.65$\pm$1.27} (1)\\
& \textbf{DeBERTa-Large}
&37.07$\pm$0.12 (5)&33.16$\pm$2.11 (7)&36.08$\pm$0.06 (6)&38.61$\pm$1.42 (4)&42.48$\pm$0.39 (3)&\textbf{46.06$\pm$0.44} (1)&\underline{45.65$\pm$0.72} (2)\\
\midrule

\multirow{4}{*}{\textbf{Senate}}
& \textbf{Skip-Gram}
&75.79$\pm$0.34 (4)&75.57$\pm$0.17 (6)&75.61$\pm$0.23 (5)&75.57$\pm$0.17 (7)&75.83$\pm$0.54 (3)&\textbf{78.33$\pm$0.98} (1)&\underline{77.22$\pm$0.25} (2)\\
& \textbf{BERT-Tiny}
&81.56$\pm$1.47 (6)&76.86$\pm$1.17 (7)&81.62$\pm$0.36 (5)&81.93$\pm$0.23 (4)&82.44$\pm$0.28 (3)&\underline{85.64$\pm$0.66} (2)&\textbf{87.13$\pm$2.70} (1)\\
& \textbf{BERT-Base}
&81.60$\pm$1.62 (6)&78.00$\pm$1.48 (7)&81.91$\pm$0.22 (4)&81.79$\pm$0.18 (5)&82.41$\pm$0.22 (3)&\underline{84.35$\pm$0.15} (2)&\textbf{85.09$\pm$2.25} (1)\\
& \textbf{DeBERTa-Large}
&82.02$\pm$1.81 (5)&77.29$\pm$2.39 (7)&82.32$\pm$0.18 (3)&81.84$\pm$0.32 (6)&\underline{82.65$\pm$0.33} (2)&82.22$\pm$1.37 (4)&\textbf{83.14$\pm$0.10} (1)\\
\midrule

\multirow{4}{*}{\textbf{House}}
& \textbf{Skip-Gram}
&78.60$\pm$1.37 (4)&76.31$\pm$0.23 (7)&78.71$\pm$2.07 (3)&78.30$\pm$0.25 (6)&78.52$\pm$1.19 (5)&\textbf{79.79$\pm$0.61} (1)&\underline{79.22$\pm$0.31} (2)\\
& \textbf{BERT-Tiny}
&82.09$\pm$1.88 (6)&79.67$\pm$1.91 (7)&83.06$\pm$0.27 (5)&83.77$\pm$0.24 (4)&86.43$\pm$0.40 (3)&\underline{87.75$\pm$0.22} (2)&\textbf{88.91$\pm$1.72} (1)\\
& \textbf{BERT-Base}
&83.60$\pm$1.93 (5)&81.15$\pm$2.54 (7)&83.43$\pm$0.19 (6)&83.67$\pm$0.19 (4)&\underline{86.49$\pm$0.07} (2)&85.61$\pm$0.32 (3)&\textbf{86.66$\pm$1.40} (1)\\
& \textbf{DeBERTa-Large}
&83.55$\pm$1.89 (5)&80.19$\pm$3.78 (7)&83.48$\pm$0.27 (6)&83.67$\pm$0.19 (4)&84.63$\pm$0.35 (3)&\underline{85.93$\pm$0.86} (2)&\textbf{86.38$\pm$0.23} (1)\\
\midrule\midrule

\multicolumn{2}{c||}{\textbf{Average Rank (AR) on \textsf{TAHB}}}
& 4.8 & 6.5 & 5.1 & 4.3 & 3.0 & 1.5 & 1.4 \\
\multicolumn{2}{c||}{\textbf{Average Rank (AR) reported in \cite{kim24:iclr}}}
& 10.7 & 15.6 & 12.6 & 10.4 & 9.4 & 3.4 & 1.7 \\
\bottomrule
\end{tabular*}
\vspace{-5mm}
\end{table*}

For node classification, following~\cite{wei22:nips, kim24:iclr}, we randomly split nodes into training, validation, and test sets with ratios of 1\%, 1\%, and 98\%, respectively, and report the average and standard deviation of accuracy over 5 random splits. 
For all HRL methods, node representations are first learned and then used to train an MLP classifier~\cite{kim24:iclr}.
For hyperedge prediction, following~\cite{kim24:iclr}, we randomly split hyperedges into training, validation, and test sets with ratios of 60\%, 20\%, and 20\%, respectively, and report AUROC over 5 random splits. 
Negative hyperedges are generated using the Size-Negative-Sampling strategy~\cite{pat20:pakdd}, where hyperedge sizes are sampled from the empirical size distribution of real hyperedges and nodes are randomly selected accordingly. 
Hyperedge representations are obtained from node representations via max-min pooling~\cite{yad19:neurips, kim24:iclr}.

\begin{table*}[t]
\centering
\footnotesize
\renewcommand{\arraystretch}{1.15}
\setlength{\tabcolsep}{5pt}
\caption{Hyperedge prediction accuracy (AUROC) of HRL methods under different text encoders on {\tahb}.}
\label{Tab5}
\vspace{-4mm}
\begin{tabular*}{\textwidth}{@{\extracolsep{\fill}}ll||ccccccc}
\toprule
\textbf{Dataset} & \textbf{Embedding}
& \textbf{HGNN} & \textbf{HyperGCN} & \textbf{HNHN} & \textbf{Allset} & \textbf{UniGCNII} & \textbf{TriCL} & \textbf{HypeBoy} \\
\midrule\midrule

\multirow{4}{*}{\textbf{Amazon Pet Supplies}}
& \textbf{Skip-Gram}
& 71.35$\pm$1.08 (7) & 71.47$\pm$2.69 (6) & 71.48$\pm$0.6 (5) & 72.1$\pm$1.2 (4) & 74.15$\pm$0.48 (3)
& \textbf{81.56$\pm$0.54} (1) & \underline{81.39$\pm$0.19} (2) \\
& \textbf{BERT-Tiny}
& 71.84$\pm$1.78 (6) & 72.97$\pm$0.84 (5) & 71.31$\pm$1.13 (7) & 73.26$\pm$0.31 (4) & 74.42$\pm$0.43 (3)
& \underline{94.35$\pm$0.43} (2) & \textbf{95.36$\pm$0.12} (1) \\
& \textbf{BERT-Base}
& 71.02$\pm$1.17 (7) & 73.02$\pm$1.28 (5) & 72.98$\pm$0.52 (6) & 73.27$\pm$0.53 (4) & 74.31$\pm$0.36 (3)
& \underline{92.83$\pm$0.62} (2) & \textbf{95.09$\pm$0.19} (1) \\
& \textbf{DeBERTa-Large}
& 72.92$\pm$2.21 (5) & 72.9$\pm$1.01 (6) & 72.49$\pm$0.62 (7) & 73.33$\pm$0.39 (4) & 74.31$\pm$0.36 (3)
& \underline{91.17$\pm$0.66} (2) & \textbf{94.94$\pm$0.33} (1) \\
\midrule

\multirow{4}{*}{\shortstack[l]{\textbf{Amazon Arts \& Crafts}}}
& \textbf{Skip-Gram}
& 62.9$\pm$0.92 (4) & 62.89$\pm$0.5 (5) & 61.34$\pm$0.35 (7) & 62.26$\pm$0.42 (6) & 64.43$\pm$0.75 (3)
& \underline{65.72$\pm$0.07} (2) & \textbf{66.84$\pm$0.2} (1) \\
& \textbf{BERT-Tiny}
& 83.0$\pm$1.85 (7) & 85.1$\pm$1.02 (3) & 84.92$\pm$1.05 (5) & 84.59$\pm$0.85 (6) & 85.09$\pm$3.11 (4)
& \underline{91.9$\pm$0.64} (2) & \textbf{93.76$\pm$0.67} (1) \\
& \textbf{BERT-Base}
& 79.76$\pm$1.07 (5) & 79.46$\pm$0.34 (6) & 78.93$\pm$1.29 (7) & 82.51$\pm$0.58 (4) & 84.43$\pm$1.64 (3)
& \underline{89.75$\pm$0.68} (2) & \textbf{92.73$\pm$1.17} (1) \\
& \textbf{DeBERTa-Large}
& 78.66$\pm$0.98 (5) & 77.98$\pm$1.13 (6) & 77.93$\pm$0.72 (7) & 82.72$\pm$0.35 (4) & 86.39$\pm$0.29 (3)
& \underline{86.4$\pm$3.65} (2) & \textbf{91.88$\pm$0.11} (1) \\
\midrule

\multirow{4}{*}{\shortstack[l]{\textbf{Amazon Grocery}}}
& \textbf{Skip-Gram}
& 63.23$\pm$1.83 (6) & 63.27$\pm$0.49 (5) & 62.31$\pm$0.32 (7) & 63.38$\pm$0.25 (4) & 64.86$\pm$0.79 (3)
& \underline{81.26$\pm$0.41} (2) & \textbf{84.78$\pm$0.12} (1) \\
& \textbf{BERT-Tiny}
& 84.7$\pm$2.29 (5) & 82.42$\pm$0.47 (7) & 84.01$\pm$1.02 (6) & 84.85$\pm$0.7 (4) & 85.75$\pm$0.8 (3)
& \underline{92.04$\pm$0.32} (2) & \textbf{95.06$\pm$0.16} (1) \\
& \textbf{BERT-Base}
& 83.69$\pm$2.16 (7) & 83.73$\pm$0.42 (6) & 83.87$\pm$0.75 (5) & 84.8$\pm$0.51 (3) & 83.94$\pm$1.98 (4)
& \underline{91.79$\pm$1.5} (2) & \textbf{95.1$\pm$0.13} (1) \\
& \textbf{DeBERTa-Large}
& 82.77$\pm$3.83 (6) & 82.92$\pm$1.05 (5) & 82.52$\pm$0.41 (7) & 83.65$\pm$0.79 (3) & 83.49$\pm$1.79 (4)
& \underline{90.52$\pm$3.02} (2) & \textbf{94.94$\pm$0.13} (1) \\
\midrule

\multirow{4}{*}{\textbf{Amazon Electronics}}
& \textbf{Skip-Gram}
& 62.94$\pm$0.9 (7) & 63.88$\pm$0.35 (4) & 63.54$\pm$0.44 (5) & 63.96$\pm$0.51 (3) & 63.31$\pm$2.06 (6)
& \underline{74.79$\pm$0.28} (2) & \textbf{75.01$\pm$0.4} (1) \\
& \textbf{BERT-Tiny}
& 74.09$\pm$0.96 (6) & 73.16$\pm$0.31 (7) & 74.43$\pm$0.59 (4) & 74.18$\pm$0.63 (5) & 75.25$\pm$2.15 (3)
& \underline{93.93$\pm$0.29} (2) & \textbf{96.09$\pm$0.2} (1) \\
& \textbf{BERT-Base}
& 82.36$\pm$0.71 (5) & 80.9$\pm$1.0 (7) & 82.3$\pm$0.81 (6) & 82.95$\pm$0.68 (3) & 82.56$\pm$2.8 (4)
& \underline{93.5$\pm$0.29} (2) & \textbf{94.75$\pm$0.27} (1) \\
& \textbf{DeBERTa-Large}
& 77.29$\pm$1.31 (5) & 77.08$\pm$2.78 (6) & 76.31$\pm$1.79 (7) & 77.97$\pm$1.29 (4) & 79.03$\pm$1.58 (3)
& \underline{91.35$\pm$3.28} (2) & \textbf{96.1$\pm$0.2} (1) \\
\midrule

\multirow{4}{*}{\textbf{arXiv Math}}
& \textbf{Skip-Gram}
& 73.32$\pm$1.44 (7) & 73.98$\pm$1.42 (5) & 73.96$\pm$1.2 (6) & 74.18$\pm$1.43 (4) & 75.24$\pm$1.26 (3)
& \textbf{84.05$\pm$0.95} (1) & \underline{83.73$\pm$1.74} (2) \\
& \textbf{BERT-Tiny}
& 86.23$\pm$1.93 (6) & 84.95$\pm$1.99 (7) & 86.44$\pm$0.63 (5) & 86.57$\pm$0.59 (4) & 87.34$\pm$0.61 (3)
& \textbf{97.32$\pm$0.6} (1) & \underline{96.19$\pm$1.15} (2) \\
& \textbf{BERT-Base}
& 86.4$\pm$1.12 (6) & 85.98$\pm$0.86 (7) & 86.53$\pm$0.87 (5) & 86.63$\pm$0.84 (4) & 87.47$\pm$0.87 (3)
& \textbf{94.45$\pm$0.63} (1) & \underline{94.04$\pm$0.83} (2) \\
& \textbf{DeBERTa-Large}
& 86.44$\pm$1.32 (7) & 86.5$\pm$0.91 (6) & 86.79$\pm$0.98 (4) & 86.55$\pm$0.8 (5) & 87.3$\pm$0.81 (3)
& \textbf{91.0$\pm$0.67} (1) & \underline{89.53$\pm$1.19} (2) \\
\midrule

\multirow{4}{*}{\textbf{arXiv Physics}}
& \textbf{Skip-Gram}
& 64.8$\pm$0.88 (6) & 56.29$\pm$0.53 (7) & 64.9$\pm$0.28 (5) & 65.31$\pm$0.88 (4) & 66.73$\pm$0.17 (3)
& \underline{73.59$\pm$0.21} (2) & \textbf{74.82$\pm$3.15} (1) \\
& \textbf{BERT-Tiny}
& 72.83$\pm$1.77 (5) & 64.36$\pm$1.23 (7) & 72.4$\pm$0.69 (6) & 74.18$\pm$0.63 (4) & 76.18$\pm$0.4 (3)
& \underline{96.89$\pm$0.25} (2) & \textbf{97.87$\pm$0.07} (1) \\
& \textbf{BERT-Base}
& 71.06$\pm$1.84 (6) & 63.93$\pm$1.15 (7) & 71.74$\pm$1.11 (5) & 73.25$\pm$0.28 (4) & 76.81$\pm$0.6 (3)
& \underline{95.27$\pm$0.38} (2) & \textbf{96.47$\pm$0.29} (1) \\
& \textbf{DeBERTa-Large}
& 67.62$\pm$1.47 (6) & 64.03$\pm$1.1 (7) & 67.74$\pm$1.9 (5) & 73.27$\pm$0.51 (4) & 77.04$\pm$0.65 (3)
& \underline{94.33$\pm$0.1} (2) & \textbf{95.11$\pm$0.27} (1) \\
\midrule

\multirow{4}{*}{\textbf{arXiv CS}}
& \textbf{Skip-Gram}
& 62.07$\pm$0.74 (5) & 57.06$\pm$2.04 (7) & 61.64$\pm$0.68 (6) & 62.42$\pm$0.45 (4) & 62.71$\pm$1.48 (3)
& \underline{67.35$\pm$0.51} (2) & \textbf{67.67$\pm$0.4} (1) \\
& \textbf{BERT-Tiny}
& 70.88$\pm$2.13 (5) & 64.74$\pm$2.66 (7) & 70.5$\pm$1.72 (6) & 72.82$\pm$0.63 (4) & 76.46$\pm$0.48 (3)
& \underline{89.42$\pm$0.24} (2) & \textbf{91.09$\pm$0.52} (1) \\
& \textbf{BERT-Base}
& 70.02$\pm$2.19 (5) & 64.57$\pm$0.62 (7) & 69.1$\pm$0.6 (6) & 72.92$\pm$0.47 (4) & 77.97$\pm$0.91 (3)
& \underline{85.96$\pm$1.09} (2) & \textbf{93.99$\pm$0.28} (1) \\
& \textbf{DeBERTa-Large}
& 70.84$\pm$2.95 (6) & 64.06$\pm$0.81 (7) & 70.96$\pm$2.9 (5) & 72.91$\pm$0.51 (4) & 76.98$\pm$1.89 (3)
& \underline{85.71$\pm$0.2} (2) & \textbf{86.99$\pm$2.07} (1) \\
\midrule

\multirow{4}{*}{\textbf{TMDB}}
& \textbf{Skip-Gram}
& 72.34$\pm$1.96 (7) & 72.75$\pm$0.32 (6) & 73.33$\pm$1.59 (5) & 81.06$\pm$0.3 (4) & 82.06$\pm$0.77 (3)
& \textbf{85.91$\pm$0.43} (1) & \underline{85.0$\pm$0.43} (2) \\
& \textbf{BERT-Tiny}
& 77.42$\pm$1.18 (6) & 76.53$\pm$1.76 (7) & 77.78$\pm$1.64 (5) & 82.84$\pm$0.27 (4) & 84.56$\pm$0.64 (3)
& \underline{92.58$\pm$0.47} (2) & \textbf{94.07$\pm$0.23} (1) \\
& \textbf{BERT-Base}
& 82.57$\pm$1.25 (5) & 76.56$\pm$1.84 (7) & 82.29$\pm$0.39 (6) & 82.78$\pm$0.52 (4) & 84.39$\pm$0.55 (3)
& \underline{90.21$\pm$2.35} (2) & \textbf{94.73$\pm$0.79} (1) \\
& \textbf{DeBERTa-Large}
& 86.92$\pm$1.18 (5) & 83.33$\pm$1.54 (7) & 84.83$\pm$0.68 (6) & 87.95$\pm$0.46 (4) & 88.59$\pm$0.82 (3)
& \underline{91.75$\pm$0.49} (2) & \textbf{92.38$\pm$0.83} (1) \\
\midrule

\multirow{4}{*}{\textbf{Senate}}
& \textbf{Skip-Gram}
& 74.48$\pm$3.15 (5) & 72.32$\pm$2.25 (7) & 75.14$\pm$0.49 (4) & 74.06$\pm$3.11 (6) & 76.43$\pm$0.42 (3)
& \textbf{84.97$\pm$0.35} (1) & \underline{82.24$\pm$1.44} (2) \\
& \textbf{BERT-Tiny}
& 79.7$\pm$0.94 (5) & 76.65$\pm$2.71 (6) & 81.29$\pm$1.67 (4) & 76.17$\pm$2.65 (7) & 82.39$\pm$0.9 (3)
& \textbf{97.66$\pm$0.3} (1) & \underline{96.22$\pm$1.21} (2) \\
& \textbf{BERT-Base}
& 75.25$\pm$2.95 (6) & 77.04$\pm$2.25 (5) & 80.65$\pm$1.73 (4) & 74.15$\pm$1.72 (7) & 81.24$\pm$0.42 (3)
& \textbf{96.97$\pm$0.06} (1) & \underline{95.85$\pm$0.17} (2) \\
& \textbf{DeBERTa-Large}
& 76.25$\pm$1.95 (6) & 69.22$\pm$2.7 (7) & 76.74$\pm$1.95 (5) & 77.92$\pm$1.08 (4) & 80.36$\pm$0.51(3)
& \textbf{94.69$\pm$0.26} (1) & \underline{93.97$\pm$1.0} (2) \\
\midrule

\multirow{4}{*}{\textbf{House}}
& \textbf{Skip-Gram}
& 62.09$\pm$1.97 (6) & 64.4$\pm$2.07 (4) & 61.04$\pm$1.2 (7) & 63.69$\pm$2.93 (5) & 66.2$\pm$2.11 (3)
& \textbf{73.47$\pm$0.24} (1) & \underline{72.95$\pm$1.61} (2) \\
& \textbf{BERT-Tiny}
& 72.58$\pm$1.67 (6) & 74.52$\pm$2.18 (4) & 71.62$\pm$2.54 (7) & 73.57$\pm$2.27 (5) & 74.89$\pm$2.57 (3)
& \underline{99.8$\pm$0.13} (2) & \textbf{99.85$\pm$0.07} (1) \\
& \textbf{BERT-Base}
& 73.14$\pm$1.43 (6) & 73.03$\pm$2.02 (7) & 73.55$\pm$2.26 (5) & 73.57$\pm$2.28 (4) & 74.69$\pm$2.63 (3)
& \textbf{99.97$\pm$0.02} (1) & \underline{99.95$\pm$0.05} (2) \\
& \textbf{DeBERTa-Large}
& 65.7$\pm$1.28 (6) & 66.05$\pm$1.73 (5) & 64.57$\pm$2.26 (7) & 66.83$\pm$1.9 (4) & 67.16$\pm$2.77 (3)
& \underline{96.92$\pm$1.35} (2) & \textbf{99.97$\pm$0.02} (1) \\
\midrule\midrule

\multicolumn{2}{c||}{\textbf{Average Rank (AR)}} 
& 5.8 & 6 & 5.6 & 4.4 & 3.2 & 1.7 & 1.3 \\
\bottomrule
\end{tabular*}
\vspace{-3mm}
\end{table*}

In addition, to analyze the impact of input features, we extract text-based input features using four categories of node encoders: the traditional shallow embedding method Skip-Gram and three pretrained language models of different scales, following the setup of~\cite{yan23:neurips}: Small (BERT-Tiny), Medium (BERT-Base), and Large (DeBERTa-Large).

Table~3 reports the node classification results on the {\tahb} datasets. 
The main observations are as follows. 
First, performance generally improves as the PLM scale increases from Small to Medium, owing to the richer contextual semantics captured by larger language models. 
However, scaling further to the Large model (DeBERTa-Large) does not consistently improve performance and occasionally leads to slight degradation. 
This observation is consistent with the findings of~\cite{yan23:neurips} and may be attributed to overfitting or an alignment gap between the high-dimensional embedding space generated by large PLMs (1024 dimensions) and the topology-centric message-passing mechanisms of hypergraph models.

Second, the relative performance trends among HRL methods are consistently reproduced on \tahb\, similar to those observed on existing hypergraph benchmark datasets. 
Specifically, the Pearson correlation coefficient (PCC) between the node classification ranking of HRL methods reported on existing hypergraph benchmarks in~\cite{kim24:iclr} and the ranking measured on \tahb\ yields a PCC of 0.97. 
This suggests that the relative performance trends validated on existing benchmarks are stably preserved in {\tahb}.\footnote{For hyperedge prediction, direct correlation analysis is difficult because no prior study evaluates all HRL methods under a unified experimental setting on existing datasets.}

Finally, we note that on large-scale datasets (\ie, arXiv Physics and arXiv CS), self-supervised HRL methods such as TriCL and HypeBoy encountered Out-of-Memory (OOM) limitations. This is primarily due to the quadratic complexity of computing high-order contrastive losses or reconstructing dense neighbor-co-membership matrices over tens of thousands of nodes. Importantly, such empirical limitations were rarely observed in existing hypergraph benchmarks due to their constrained data scales. By exposing these hidden computational bottlenecks for the first time, {\tahb} uniquely serves as a critical testbed for evaluating and advancing scalability in high-order representation learning, paving the way for future memory-efficient frameworks.

Table~4 reports the hyperedge prediction results on the {\tahb} datasets. 
Overall, the observed performance trends are largely consistent with those of node classification. 
These results demonstrate that {\tahb} faithfully preserves the task difficulty and topological discriminability of existing hypergraph benchmarks.

\vspace{1mm}
\noindent \textbf{Summary of Findings.}
Overall, the experimental results in this section demonstrate that {\tahb} not only faithfully preserves the structural characteristics of real-world hypergraphs, but also effectively retains the distributional and semantic properties of textual attributes observed in real-world text-attributed data. 
Moreover, the performance trends of various hypergraph representation learning models and text encoding methods observed on existing benchmark datasets are consistently reproduced on {\tahb}. 
These findings suggest that {\tahb} can serve as a reliable and scalable benchmark for next-generation text-aware hypergraph representation learning that jointly leverages structural and textual information.

\vspace{1mm}
\noindent \textbf{Ethics, Privacy, and Potential Risks}. To ensure the highest standard of research ethics, all benchmark datasets underlying {\tahb} were curated from public repositories under appropriate creative commons or scientific data licenses. A paramount priority during our preprocessing phase was safeguarding user privacy; hence, we executed text anonymization protocols to strip away all personal identifiers, account names, and specific location markers. Regarding potential downstream risks, web-scraped content inevitably conveys historical or societal biases latent within user-generated text. However, because {\tahb} functions solely as a topological and semantic benchmark for machine learning research, its utility lies in training structural encoders rather than text generation. We actively encourage the research community to utilize {\tahb} within fair, safe, and aligned AI practices.
\section{Exploring LLM Integration in Hypergraph Learning }~\label{s4}

\vspace{-3mm}
\noindent \textbf{Motivation.}
Existing studies on integrating large language models (LLMs) with graph learning have primarily focused on standard graph settings, while their applicability to hypergraph learning remains largely unexplored. 
Using {\tahb}, which jointly provides real-world hypergraph structures and raw textual attributes, we present the first systematic analysis of the integration between LLMs and hypergraph representation learning (HRL). 
Specifically, we investigate this integration from two complementary perspectives: \textit{LLM-as-Predictor}, where LLMs directly perform downstream tasks, and \textit{LLM-as-Enhancer}, where LLMs are utilized as knowledge augmentation modules for HRL.

\noindent \textbf{LLM-as-Predictor.}
LLM-as-Predictor refers to a setting in which LLMs directly perform downstream tasks as standalone prediction models. 
Unlike conventional hypergraph neural network-based methods that explicitly learn structural information through message passing, this setting converts both hypergraph structural information and textual attributes into natural language prompts and directly feeds them into LLMs. 
Based on the provided prompts, the LLM performs downstream prediction tasks.

In this section, we investigate whether LLMs can effectively utilize not only textual semantics but also high-order topology induced by hyperedge co-membership, thereby exploring the potential of LLM--hypergraph integration. 
Specifically, neighboring information derived from the nodes belonging to the same hyperedge is transformed into textual prompts and incorporated into the LLM input.
We conduct node classification as a downstream task.\footnote{For hyperedge prediction, all positive and negative candidate hyperedges should be individually converted into LLM inputs, resulting in \textit{prohibitively high monetary cost}. Therefore, we only perform node classification under the LLM-as-Predictor setting.} 
To analyze the impact of structural and textual information on hypergraph learning, we consider three input settings: (1) \textit{Topology-only}, only the hyperedge membership (\ie, a set of neighboring nodes) information associated with the target node is converted into natural language prompts; (2) \textit{Text-only}, where only the textual attributes of the target node are provided; and (3) \textit{Topology+Text}, where both textual attributes and hypergraph structural information are provided.
For all three settings, we use the same prompt template and inference protocol, and measure the final prediction accuracy using the class label generated by the LLM. 
We evaluate three representative commercial and open-source LLMs: ChatGPT-4o, Llama-3.1, and Gemini-2.5.
Finally, due to the high monetary cost of LLM inference, we follow~\cite{che23:arxiv} by randomly sampling 15\% of the hyperedges from each dataset to construct a sub-hypergraph, on which node classification is performed.

Table~5 reports the node classification results under the three prompt settings.
The main observations are summarized as follows.
First, across all datasets and regardless of the LLM used (ChatGPT-4o, Llama-3.1, and Gemini-2.5), the \textit{Topology+Text} setting consistently achieves the best performance. 
This suggests that textual semantics and hypergraph topology provide complementary information, and that jointly leveraging both enables the most effective prediction performance.
Second, the \textit{Text-only} setting consistently outperforms the \textit{Topology-only} setting. 
This observation suggests that LLMs, being fundamentally pretrained on natural language corpora, are inherently more effective at understanding and utilizing textual semantics.
In contrast, leveraging hypergraph topology requires LLMs to infer structural relationships among multiple nodes connected through shared hyperedges, which goes beyond simple semantic extraction from textual information. 
In other words, topology-based prediction requires relatively more-complex multi-step structural reasoning than textual-semantic prediction, which may explain the comparatively limited performance of current general-purpose LLMs under the \textit{Topology-only} setting.

Overall, the experimental results suggest that current general-purpose LLMs can effectively utilize not only textual semantics but also high-order topology induced by hyperedge co-membership as complementary structural information for downstream tasks. 
However, the consistently lower performance of the \textit{Topology-only} setting compared with the \textit{Text-only} setting indicates that, while current LLMs exhibit strong capability in textual reasoning, they may still face limitations in complex hypergraph structural reasoning.

\begin{table*}[t]
\centering
\footnotesize

\setlength{\tabcolsep}{3pt}
\renewcommand{\arraystretch}{1.15}
\setlength{\aboverulesep}{2pt}
\setlength{\belowrulesep}{2pt}
\caption{Node classification accuracy of LLM-as-Predictor under different prompts and LLMs on {\tahb}.}
\vspace{-4mm}
\label{Tab4}

\begin{tabular*}{\textwidth}{@{\extracolsep{\fill}}llcccccccccc}
\toprule
\textbf{LLM} & \textbf{Prompt} 
& \makecell[c]{\textbf{Amazon}\\\textbf{Pet Supplies}}
& \makecell[c]{\textbf{Amazon}\\\textbf{Arts \& Crafts}}
& \makecell[c]{\textbf{Amazon}\\\textbf{Grocery}}
& \makecell[c]{\textbf{Amazon} \\ \textbf{Electronics}}
& \makecell[c]{\textbf{arXiv}\\\textbf{Math}}
& \makecell[c]{\textbf{arXiv}\\\textbf{Physics}}
& \makecell[c]{\textbf{arXiv} \\ \textbf{CS}}
& \textbf{TMDB}
& \textbf{Senate} 
& \textbf{House} \\
\midrule
\midrule

\multirow{3}{*}{\textbf{GPT}}
& \textbf{Topology Only}  
& 57.80$\pm$1.68
& 57.20$\pm$0.19
& 51.63$\pm$1.38
& 56.40$\pm$4.52
& 81.40$\pm$3.55
& 80.47$\pm$2.54
& 68.63$\pm$1.38
& 26.74$\pm$0.84
& 72.00$\pm$2.24
& \underline{75.93$\pm$2.81} \\

& \textbf{Text Only}    
& \underline{65.93$\pm$0.14}
& \underline{70.93$\pm$0.68}
& \underline{58.11$\pm$1.32}
& \underline{70.18$\pm$1.24}
& \underline{82.27$\pm$1.38}
& \underline{87.27$\pm$0.01}
& \underline{88.11$\pm$1.32}
& \underline{46.96$\pm$2.51}
& \underline{73.27$\pm$3.56}
& 74.47$\pm$1.53 \\

& \textbf{Topology+Text}  
& \textbf{68.73$\pm$0.25}
& \textbf{74.67$\pm$0.70}
& \textbf{58.66$\pm$0.60}
& \textbf{73.80$\pm$0.13}
& \textbf{82.53$\pm$0.12}
& \textbf{87.60$\pm$1.04}
& \textbf{88.66$\pm$0.60}
& \textbf{50.14$\pm$1.76}
& \textbf{76.20$\pm$1.31}
& \textbf{79.20$\pm$2.43} \\

\midrule

\multirow{3}{*}{\textbf{LLAMA}}
& \textbf{Topology Only}  
& 25.40$\pm$0.32
& 50.47$\pm$0.28
& 51.20$\pm$0.11
& 22.90$\pm$1.23
& 17.40$\pm$1.23
& 63.27$\pm$3.05
& 26.06$\pm$0.67
& 0.98$\pm$0.21
& 62.93$\pm$0.17
& 64.40$\pm$1.71 \\

& \textbf{Text Only}   
& \underline{65.67$\pm$1.61}
& \underline{65.40$\pm$2.04}
& \underline{51.87$\pm$1.74}
& \underline{72.02$\pm$2.76}
& \underline{40.27$\pm$1.72}
& \underline{77.93$\pm$4.52}
& \underline{74.72$\pm$0.73}
& \underline{41.32$\pm$2.57}
& \underline{64.93$\pm$3.24}
& \textbf{66.80$\pm$2.84} \\

& \textbf{Topology+Text}  
& \textbf{66.20$\pm$0.99}
& \textbf{68.93$\pm$3.90}
& \textbf{55.20$\pm$0.08}
& \textbf{74.64$\pm$2.11}
& \textbf{54.40$\pm$0.51}
& \textbf{80.00$\pm$1.15}
& \textbf{80.30$\pm$1.06}
& \textbf{45.10$\pm$0.87}
& \textbf{65.07$\pm$2.30}
& \underline{66.73$\pm$1.18} \\

\midrule

\multirow{3}{*}{\textbf{Gemini}}
& \textbf{Topology Only}  
& 54.72$\pm$2.54
& 54.64$\pm$3.18
& 51.36$\pm$2.93
& 51.66$\pm$1.06
& 80.48$\pm$0.54
& 80.40$\pm$0.52
& 69.16$\pm$1.03
& 24.08$\pm$1.81
& 64.22$\pm$2.03
& 66.00$\pm$1.54 \\

& \textbf{Text Only}    
& \underline{63.02$\pm$1.47}
& \underline{70.96$\pm$1.62}
& \underline{53.32$\pm$1.80}
& \underline{65.70$\pm$1.47}
& \underline{85.16$\pm$0.19}
& \underline{84.66$\pm$0.31}
& \underline{87.46$\pm$0.75}
& \underline{48.02$\pm$1.02}
& \underline{68.40$\pm$1.74}
& \underline{70.70$\pm$1.26} \\

& \textbf{Topology+Text}   
& \textbf{68.82$\pm$2.12}
& \textbf{73.70$\pm$0.49}
& \textbf{56.86$\pm$0.79}
& \textbf{68.24$\pm$2.65}
& \textbf{86.94$\pm$0.47}
& \textbf{87.88$\pm$0.05}
& \textbf{89.96$\pm$0.82}
& \textbf{49.38$\pm$3.06}
& \textbf{70.44$\pm$1.36}
& \textbf{75.96$\pm$0.80} \\

\bottomrule
\end{tabular*}
\vspace{-5mm}
\end{table*}
\begin{table*}[t]
\centering
\footnotesize

\setlength{\tabcolsep}{3pt}
\renewcommand{\arraystretch}{1.15}
\setlength{\aboverulesep}{2pt}
\setlength{\belowrulesep}{2pt}
\caption{Node classification accuracy of HGNN with LLM-based textual augmentation on {\tahb}.}
\label{Tab3}
\vspace{-4mm}
\begin{tabular*}{\textwidth}{@{\extracolsep{\fill}}llcccccccccc}
\toprule
\textbf{Methods} 
& \textbf{LLM} 
& \makecell[c]{\textbf{Amazon}\\\textbf{Pet Supplies}}
& \makecell[c]{\textbf{Amazon}\\\textbf{Arts \& Crafts}}
& \makecell[c]{\textbf{Amazon}\\\textbf{Grocery}}
& \makecell[c]{\textbf{Amazon} \\ \textbf{Electronics}}
& \makecell[c]{\textbf{arXiv}\\\textbf{Math}}
& \makecell[c]{\textbf{arXiv}\\\textbf{Physics}}
& \makecell[c]{\textbf{arXiv} \\ \textbf{CS}}
& \textbf{TMDB}
& \textbf{Senate} 
& \textbf{House} \\
\midrule
\midrule

\textbf{Original Text} & --
& 81.97$\pm$0.43
& 75.78$\pm$0.93
& 68.23$\pm$3.66
& 71.47$\pm$0.91
& 72.28$\pm$2.56
& 95.06$\pm$0.16
& 81.19$\pm$0.22
& 35.35$\pm$2.16
& 81.56$\pm$1.47
& 82.09$\pm$1.88 \\

\midrule

\multirow[c]{3}{*}{\textbf{Enhanced Text}}
& \textbf{GPT}
& \textbf{84.93$\pm$0.17}
& \textbf{80.15$\pm$0.52}
& \underline{70.09$\pm$3.89}
& \textbf{74.68$\pm$0.58}
& \underline{74.16$\pm$2.58}
& \textbf{96.66$\pm$0.49}
& \textbf{86.14$\pm$0.01}
& \textbf{39.61$\pm$1.81}
& \textbf{86.48$\pm$1.32}
& \underline{84.72$\pm$1.67} \\

& \textbf{LLAMA}
& \underline{84.17$\pm$0.87}
& \underline{78.86$\pm$1.10}
& \textbf{70.59$\pm$3.26}
& \underline{73.93$\pm$0.88}
& \textbf{74.74$\pm$2.90}
& 96.11$\pm$0.01
& 84.32$\pm$0.01
& 39.13$\pm$2.60
& 84.73$\pm$1.12
& \textbf{84.94$\pm$1.52} \\

& \textbf{Gemini}
& 83.71$\pm$0.29
& 78.29$\pm$0.67
& 68.90$\pm$3.85
& 72.48$\pm$0.49
& 74.09$\pm$2.83
& \underline{96.19$\pm$0.61}
& \underline{85.20$\pm$0.30}
& \underline{39.33$\pm$1.79}
& \underline{85.22$\pm$1.55}
& 83.20$\pm$2.35 \\

\bottomrule
\end{tabular*}
\vspace{-4mm}
\end{table*}

\vspace{1mm}
\noindent \textbf{LLM-as-Enhancer.}
LLM-as-Enhancer refers to a setting in which LLMs are not used as standalone prediction models, but instead serve as semantic augmentation modules that enhance the input features of existing hypergraph representation learning models. 
In this section, we adapt an LLM-based textual augmentation strategy originally proposed for graph learning~\cite{he24:iclr} to hypergraph environments, thereby exploring another potential direction for LLM--hypergraph integration. 
The key idea is that LLM-generated reasoning can enrich raw textual attributes with implicit task-relevant semantic information beyond what is explicitly expressed in the original text.

Specifically, the textual attributes of each node are first provided to the LLM, which is then prompted to generate both the most likely class label and the corresponding reasoning for the target node. 
The generated reasoning is subsequently utilized as an augmented textual attribute enriched with additional semantic information.
Next, both the original textual attributes and the LLM-generated augmented textual attributes are jointly fed into PLMs to extract node features. 
Finally, the extracted node features are used as input features for existing hypergraph neural networks to perform downstream tasks.

To investigate the applicability of LLM-based textual augmentation in hypergraph learning, we conduct node classification as the downstream task.\footnote{Most existing LLM-based graph learning methods have primarily focused on node classification tasks~\cite{he24:iclr, jin24:tkde}. 
Accordingly, these methods are largely optimized for node classification settings, while their extension to link prediction remains relatively underexplored. 
We leave the investigation of LLM-based hyperedge prediction as an important direction for future research.} 
We use BERT-Tiny as the PLM encoder and HGNN as the backbone HRL model. 
All experiments follow the same evaluation protocols described in Section~3.3-VQ3.
Table~6 reports the node classification results using LLM-based textual augmentation under different LLMs.
The main observations are summarized as follows. 
First, applying LLM-based textual augmentation consistently improves performance compared with using only the original textual attributes. 
This suggests that the reasoning-generated text produced by LLMs can provide richer semantic information than raw textual attributes alone.
Second, the effectiveness of textual augmentation generally follows the order of ChatGPT-4o, Llama-3.1, and Gemini-2.5. 
This observation indicates that the reasoning quality and semantic generation capability of LLMs can directly influence the final performance of hypergraph representation learning models.
Overall, these results demonstrate that the LLM-based textual augmentation originally proposed in graph learning environments can also be effectively applied to hypergraph settings, suggesting a promising direction for future LLM--hypergraph integration research.

\vspace{-3mm}
\section{Future Research Directions}~\label{s5}
{\tahb} extends beyond a benchmark for evaluating existing hypergraph representation learning (HRL) methods and opens up diverse future research directions at the intersection of LLMs and hypergraph learning. 
By jointly providing real-world hypergraph structures and raw textual attributes, {\tahb} establishes a foundation for systematically exploring research problems that are difficult to investigate in conventional topology-centric HRL settings.

\vspace{1mm}
\noindent \textbf{Beyond Node-Level Downstream Tasks.}
Most existing text-aware graph learning studies primarily focus on node-level tasks such as node classification. 
However, real-world hypergraphs naturally involve higher-level downstream tasks, including hyperedge-level and hypergraph-level prediction problems. 
Future research directions include hyperedge prediction, hyperedge classification, group recommendation, and whole-hypergraph classification by jointly leveraging textual semantics and high-order topology. 
By explicitly modeling groupwise interactions among multiple nodes, {\tahb} provides a realistic benchmark for unified node-, hyperedge-, and hypergraph-level learning frameworks.

\vspace{1mm}
\noindent \textbf{LLM-based Hypergraph Reasoning.}
Our experimental results suggest that, while current general-purpose LLMs can effectively leverage textual semantics, they still face substantial challenges in complex hypergraph structural reasoning. 
In particular, the consistently lower performance of topology-only settings compared with text-only settings indicates that understanding and reasoning over high-order topology induced by hyperedge co-membership remains challenging for existing LLMs. 
This observation opens up promising future directions such as hypergraph-aware prompting, structural chain-of-thought reasoning, and hyperedge-level reasoning agents. 
By providing realistic text-attributed hypergraphs, {\tahb} can serve as a foundational benchmark for evaluating LLM-based hypergraph reasoning frameworks.

\vspace{1mm}
\noindent \textbf{Text-aware Hypergraph Learning.}
Existing HRL research has primarily focused on topology-driven learning, while studies jointly considering textual semantics remain relatively limited. 
However, our results suggest that semantic information and structural information can provide complementary signals in text-aware hypergraph environments. 
{\tahb} enables a broad range of future research directions, including semantic-aware hypergraph representation learning, text-guided hyperedge modeling, and semantic-structural co-learning. 
In particular, {\tahb} provides a realistic evaluation environment for developing representation learning frameworks that jointly model textual semantics and high-order topology.

\vspace{1mm}
\noindent \textbf{Hypergraph Foundation Models.}
Although graph foundation models and graph-language models have recently attracted significant attention, most existing studies focus on standard graph settings, while foundation model research for hypergraphs remains at an early stage. 
A major limitation of existing hypergraph benchmarks is that they rarely provide raw textual attributes together with high-order structural information, making it difficult to study foundation models that jointly learn semantics and topology. 
By providing diverse text-attributed hypergraphs across multiple domains, {\tahb} can serve as a pretraining benchmark for future hypergraph foundation models. 
Furthermore, it can facilitate various downstream research directions such as instruction tuning, retrieval-augmented hypergraph learning, and LLM-aligned hypergraph embeddings.

\enlargethispage{2\baselineskip}
\section{Conclusions}~\label{s6}
In this paper, we introduce {\tahb}, the first publicly available benchmark specifically designed for text-aware hypergraph learning, which seamlessly bridges the gap between high-order groupwise relations and raw textual semantics. Through rigorous and extensive validation, we demonstrate that {\tahb} faithfully preserves the fundamental structural topologies of real-world hypergraphs while retaining informative textual distributions, thereby establishing a highly reliable and realistic evaluation environment. Leveraging this infrastructure, we systematically explore the integration of LLMs with hypergraph representation learning under two complementary paradigms: LLM-as-Predictor and LLM-as-Enhancer. Our empirical findings reveal that current LLMs are exceptionally powerful as semantic enhancers but still face inherent limitations in multi-step structural reasoning when operating as standalone predictors. Moving forward, {\tahb} serves as a critical testbed for uncovering empirical computational bottlenecks and scaling up high-order learning frameworks. Ultimately, we anticipate that this benchmark will catalyze future research and pave the way toward advanced hypergraph foundation models that jointly optimize topological structures and rich natural language semantics.

\section*{GenAI Usage Disclosure}~\label{s7}
In accordance with the ACM Authorship Policy, we disclose that generative AI tools were selectively employed during this research. 
Generative AI models (including ChatGPT-4o, Llama-3.1, and Gemini-2.5) were utilized to implement the LLM-as-Predictor and LLM-as-Enhancer frameworks. Except for limited writing assistance, these models were not used in any other stages of the research process, including data collection, preprocessing, experimental evaluation, or result analysis.
GenAI-assisted writing refinement was strictly limited to assist in editing and polishing the author-written content, including minor grammar correction, phrasing edits, and word-level auto correction.

\bibliographystyle{ACM-Reference-Format}
\bibliography{sample-base}

@inproceedings{jang23:cikm,
  title={SAGE: A Storage-Based Approach for Scalable and Efficient Sparse Generalized Matrix-Matrix Multiplication},
  author={M. Jang and Y. Ko and H. Gwon and I. Jo and Y. Park and S. Kim},
  booktitle={Proceedings of ACM International Conference on Information and Knowledge Management (CIKM)},
  year={2023},
}

@inproceedings{yoo23:www,
  title={Disentangling Degree-related Biases and Interest for Out-of-Distribution Generalized Directed Network Embedding},
  author={H. Yoo and Y. Lee and K. Shin and S. Kim},
    booktitle = {Proceedings of ACM Web Conference (WWW)},
  year={2023},
}

@article{new14:pnas,
  title="Coauthorship Networks and Patterns of Scientific Collaboration",
  author="M. Newman",
  journal="Proceedings National Academy of Sciences USA",
  year="2014",
}

@inproceedings{zh20:aaai,
  title="Defining and Evaluating Network Communities Based on Ground-Truth",
  author="C. Zheng and X. Fan and C. Wang and J. Qi",
  booktitle = {Proceedings of AAAI Conference on Artificial Intelligence (AAAI)},
  year= "2020",
}

@inproceedings{he21:icml,
  title={Improving Molecular Graph Neural Network Explainability with Orthonormalization and Induced Sparsity},
  author={R. Henderson and D. Clevert and F. Montanari},
  booktitle = {Proceedings of International Conference on Machine Learning (ICML)},
  year={2021},
}

@inproceedings{ya12:icdm,
  title="Defining and Evaluating Network Communities Based on Ground-Truth",
  author="J. Yang and J. Leskovec",
  booktitle="Proceedings of IEEE International Conference on Data Mining (ICDM)",
  year="2012",
}

@inproceedings{zhou06:nips,
  title={Learning with Hypergraphs: Clustering, Classification, and Embedding},
  author={D. Zhou and J. Huang and B. Scholköpf},
  booktitle = {Proceedings of Conference on Neural Information Processing Systems (NIPS)},
  year      = {2006},
}

@inproceedings{do20:kdd,
  author    = {M. Do and S. Yoon and B. Hooi and K. Shin},
  title     = {Structural Patterns and Generative Models of Real-World Hypergraphs},
  booktitle={Proceedings of ACM International Conference on Knowledge Discovery and Data Mining (KDD)},
  year      = {2020},
}

@inproceedings{zha20:iclr,
  author    = {R. Zhang and Y. Zou and J. Ma},
  title     = {Hyper-SAGNN: A Self-Attention Based Graph Neural Network for Hypergraphs},
  booktitle = {Proceedings of International Conference on Learning Representations (ICLR)},
  year      = {2020},
}

@inproceedings{han23:web,
  title={Intra and Inter Domain HyperGraph Convolutional Network for Cross-Domain Recommendation},
  author={Z. Han and X. Zheng and C. Chen and W. Cheng and Y. Yao},
    booktitle = {Proceedings of ACM Web Conference (WWW)},
  year={2023},
}

@inproceedings{yu21:www,
  title={Self-Supervised Multi-Channel Hypergraph Convolutional Network for Social Recommendation},
  author={J. Yu and H. Yin and J. Li and Q. Wang and N. Hung and X. Zhang},
    booktitle = {Proceedings of ACM Web Conference (WWW)},
  year      = {2021},
}

@inproceedings{fen19:aaai,
  author    = {Y. Feng and H. You and Z. Zhang and R. Ji and Y. Gao},
  title     = {Hypergraph Neural Networks},
  booktitle = {Proceedings of AAAI Conference on Artificial Intelligence (AAAI)},
  year      = {2019},
}

@inproceedings{chi21:iclr,
  title={You are AllSet: A Multiset Function Framework for Hypergraph Neural Networks},
  author={E. Chien and C. Pan and J. Peng and O. Milenkovic},
  booktitle = {Proceedings of International Conference on Learning Representations (ICLR)},
  year={2021},
}

@inproceedings{vel20:kdd,
  title   = {Minimizing Localized Ratio Cut Objectives in Hypergraphs},
  author  = {N. Veldt and A. Benson and J. Kleinberg},
  booktitle={Proceedings of ACM International Conference on Knowledge Discovery and Data Mining (KDD)},
  year    = {2020},
}

@inproceedings{xu2019:iclr,
  title={How Powerful are Graph Neural Networks?},
  author={K. Xu and W. Hu and J. Leskovec and S. Jegelka},
  booktitle={Proceedings of International Conference on Learning Representations (ICLR)},
  year={2019}
}

@inproceedings{zhu20:neurips,
  title={Beyond Homophily in Graph Neural Networks: Current Limitations and Effective Designs},
  author={J. Zhu and Y. Yan and L. Zhao and M. Heimann and L. Akoglu and D. Koutra},
  booktitle={Proceedings of Advances in Neural Information Processing Systems (NeurIPS)},
  year={2020}
}

@inproceedings{yang22:www,
  title={Graph Neural Networks Beyond Compromise Between Attribute and Topology},
  author={L. Yang and W. Zhou and W. Peng and B. Niu and J. Gu and C. Wang and X. Cao and D. He},
  booktitle={Proceedings of the ACM Web Conference (WWW)},
  year={2022}
}

@article{jin24:tkde,
  title={Large Language Models on Graphs: A Comprehensive Survey},
  author={B. Jin and G. Liu and C. Han and M. Jiang and H. Ji and J. Han},
  journal={IEEE Transactions on Knowledge and Data Engineering},
  year={2024},
}

@article{pen24:arxiv,
  title={Learning on Multimodal Graphs: A Survey},
  author={C. Peng and J. He and F. Xia},
  journal={arXiv preprint arXiv:2402.05322},
  year={2024}
}

@inproceedings{kan21:icdm,
  title="Adversarial Learning of Balanced Triangles for Accurate Community Detection on Signed Networks",
  author="Y. Kang and W. Lee and Y. Lee and K. Han and S. Kim",
  booktitle="Proceedings of IEEE International Conference on Data Mining (ICDM)",
  year="2021",
}

@article{spa72:jd,
  title={A Statistical Interpretation of Term Specificity and Its application in Retrieval},
  author={K. Sparck Jones},
  journal={Journal of Documentation},
  year={1972},
}

@article{mik13:arxiv,
  title={Efficient Estimation of Word Representations in Vector Space},
  author={T. Mikolov and K. Chen and G. Corrado and J. Dean},
  journal={arXiv preprint arXiv:1301.3781},
  year={2013},
}

@inproceedings{bro20:neurips,
  title={Language Models are Few-Shot Learners},
  author={T. Brown and B. Mann and N. Ryder and M. Subbiah and J. Kaplan and P. Dhariwal and A. Neelakantan and P. Shyam and G. Sastry and A. Askell and others},
  booktitle={Proceedings of Advances in Neural Information Processing Systems (NeurIPS)},
  year={2020}
}

@article{raf20:jmlr,
  title={Exploring the Limits of Transfer Learning with a Unified Text-to-Text Transformer},
  author={C. Raffel and N. Shazeer and A. Roberts and K. Lee and S. Narang and M. Matena and Y. Zhou and W. Li and P. Liu},
  journal={Journal of machine learning research},
  year={2020}
}

@article{tou23:arxiv,
  title={Llama 2: Open Foundation and Fine-tuned Chat Models},
  author={H. Touvron and L. Martin and K. Stone and P. Albert and A. Almahairi and Y. Babaei and N. Bashlykov and S. Batra and P. Bhargava and S. Bhosale and others},
  journal={arXiv preprint arXiv:2307.09288},
  year={2023}
}

@inproceedings{yan23:neurips,
  title={A Comprehensive Study on Text-Attributed Graphs: Benchmarking and Rethinking},
  author={H. Yan and C. Li and R. Long and C. Yan and J. Zhao and W. Zhuang and J. Yin and P. Zhang and W. Han and H. Sun and others},
  booktitle={Proceedings of Advances in Neural Information Processing Systems (NeurIPS)},
  year={2023}
}

@inproceedings{jin23:acl,
  title={Patton: Language Model Pretraining on Text-Rich Networks},
  author={B. Jin and W. Zhang and Y. Zhang and Y. Meng and X. Zhang and Q. Zhu and J. Han},
  booktitle={Proceedings of the Annual Meeting of the Association for Computational Linguistics (ACL)},
  year={2023}
}

@inproceedings{yad19:neurips,
  title="HyperGCN: A New Method of Training Graph Convolution Networks on Hypergraphs",
  author="N. Yadti and M. Nimishakavi and P. Yadav and V. Nitin and A. Louis and P. Talukdar",
  booktitle="Proceedings of Conf. on Neural Information Processing Systems (NeurIPS)",
  year="2019"
}

@inproceedings{wan19:www,
  title="Heterogeneous Graph Attention Network",
  author="X. Wang and H. Ji and C. Shi and B. Wang and Y. Ye and P. Cui and P. Yu",
    booktitle = "Proceedings of ACM Web Conf. (WWW)",
  year      = "2019"
}

@inproceedings{zha19:kdd,
  title={Heterogeneous Graph Neural Network},
  author={C. Zhang and D. Song and C. Huang and A. Swami and N. Chawla},
  booktitle = "Proceedings of ACM Int'l Conf. on Knowledge Discovery and Data Mining (KDD)",
  year={2019}
}

@inproceedings{wu15:cvpr,
  title="3D ShapeNets: A Deep Representation for Volumetric Shapes",
  author="Z. Wu and S. Song and A. Khosla and F. Yu and L. Zhang and X. Tang and J. Xiao",
    booktitle = "Proceedings of IEEE Conf. on Computer Vision and Pattern Recognition (CVPR)",
  year      = "2015"
}

@inproceedings{ci21:iclr,
  author    = "E. Chien and C. Pan and J. Peng and O. Milenkovic",
  title     = "You are Allset: A Multiset Function Framework for Hypergraph Neural Networks",
  booktitle = "Proceedings of Int'l Conf. on Learning Representations (ICLR)",
  year      = "2021"
}

@inproceedings{che03:cg,
  title={On Visual Similarity Based 3D Model Retrieval},
  author={D. Chen and X. Tian and Y. Shen and M. Ouhyoung},
  booktitle={Computer Graphics Forum},
  year={2003},
}

@inproceedings{sun08:kdd,
  author    = "L. Sun and S. Ji and J. Ye",
  title     = "Spectral Learning for Multi-Label Classification",
  booktitle = "Proceedings of ACM Int'l Conf. on Knowledge Discovery and Data Mining (KDD)",
  year      = "2008"
}

@inproceedings{zha24:neurips,
  title={DTGB: A Comprehensive Benchmark for Dynamic Text-Attributed Graphs},
  author={J. Zhang and J. Chen and M. Yang and A. Feng and S. Liang and J. Shao and R. Ying},
  booktitle={Proceedings of Advances in Neural Information Processing Systems (NeurIPS)},
  year={2024}
}

@inproceedings{kim24:iclr,
    author = "S. Kim and S. Kang and F. Bu and S. Lee and J. Yoo and K. Shin",
    title = "HypeBoy: Generative Self-Supervised Represenation Learning on Hypergraphs",
    booktitle = "Proc. of Int'l Conf. on Learning Representations (ICLR)",
    year = "2024" 
}

@inproceedings{pat20:pakdd,
  title={Negative Sampling for Hyperlink Prediction in Networks},
  author={P. Patil and G. Sharma and M. Murty},
  booktitle={Proc. of  Pacific-Asia Conf. on Knowledge Discovery and Data Mining (PAKDD)},
  year={2020}
}

@inproceedings{don20:icmlw,
  title         = "HNHN: Hypergraph Networks with Hyperedge Neurons",
  author 	= "Y. Dong and W. Sawin and Y. Bengio",
  booktitle 	= "ICML Graph Representation Learning and Beyond Workshop",
  year   = "2020"
 }

@inproceedings{hua21:ijcai,
    author = "J. Huang and J. Yang",
    title = "UniGNN: a Unified Framework for Graph and Hypergraph Neural Networks",
    booktitle = "Proc. of Int'l Joint Conf. on Artificial Intelligence (IJCAI)",
    year = "2021"
}

@inproceedings{lee23:aaai,
  author    = "D. Lee and K. Shin",
  title     = "I’m me, we’re us, and i’m us: Tri-directional contrastive learning on hypergraphs",
  booktitle = "Proc. of AAAI Conf. on Artificial Intelligence (AAAI)",
  year      = "2023"
}

@inproceedings{wei22:nips,
    author = "T. wei and Y. You and T. Chen and Y. Shen and J. He and Z. Wang",
    title = "Augmentations in Hypergraph Contrastive Learning: Fabricated and Generative",
    booktitle = "Proc. of the Conf. on Advances in Neural Information Processing Systems (NeurIPS)",
    year = "2022"
}

@article{che23:arxiv,
  title={Exploring the Potential of Large Language Models (LLMs) in Learning on Graphs},
  author={Z. Chen and H. Mao and H. Li and W. Jin and H. Wen and X. Wei and S. Wang and D. Yin and W. Fan and H. Liu and J. Tang},
  journal={arXiv preprint arXiv:2307.03393},
  year={2023}
}

@inproceedings{he24:iclr,
  title={Harnessing Explanations: LLM-to-LM Interpreter for Enhanced Text-Attributed Graph Representation Learning},
  author={X. He and X. Bresson and T. Laurent and A. Perold and Y. LeCun and B. Hooi},
    booktitle = "Proc. of Int'l Conf. on Learning Representations (ICLR)",
  year={2024}
}

@inproceedings{kip17:iclr,
  author    = "T. Kipf and M. Welling",
  title     = "Semi-Supervised Classification with Graph Convolutional Networks",
  booktitle = "Proc. of Int'l Conf. on Learning Representations (ICLR)",
  year      = "2017"
}

@article{vel17:arxiv,
  title="Graph Attention Networks",
  author="P. Veli{\v{c}}kovi{\'c} and G. Cucurull and A. Casanova and A. Romero and P. Lio and Y. Bengio",
  journal="arXiv preprint arXiv:1710.10903",
  year="2017"
}

@article{hou24:arxiv,
  title={Bridging Language and Items for Retrieval and Recommendation},
  author={Y. Hou and J. Li and Z. He and A. Yan and X. Chen and J. McAuley},
  journal={arXiv preprint arXiv:2403.03952},
  year={2024}
}

@inproceedings{tang09:kdd,
  title={Relational Learning via Latent Social Dimensions},
  author={L. Tang and H. Liu},
  booktitle={KDD},
  year={2009}
}

@inproceedings{hu20:neurips,
  title={Open Graph Benchmark: Datasets for Machine Learning on Graphs},
  author={W. Hu and M. Fey and M. Zitnik and Y. Dong and H. Ren and B. Liu and M. Catasta and J. Leskovec},
  booktitle={NeurIPS},
  year={2020}
}

@inproceedings{he24:neurips,
  title={TEG-DB: A Comprehensive Benchmark for Text-Enriched Graph Learning},
  author={Z. Li and Z. Gou and X. Zhang and Z. Liu and S. Li and Y. Hu and C. Ling and Z. Zhang and L. Zhao},
  booktitle={NeurIPS},
  year={2024}
}

@article{ben18:pnas,
  title={Simplicial Closure and Higher-Order Link Prediction},
  author={A. Benson and R. Abebe and M. Schaub and A. Jadabaie and J. Kleinberg},
  journal={PNAS},
  year={2018}
}

@inproceedings{neo24:cikm,
  title={Towards Fair Graph Anomaly Detection: Problem, Benchmark Datasets, and Evaluation},
  author={N. Neo and Y. Lee and Y. Jin and S. Kim and S. Kumar},
  booktitle={Proceedings of ACM International Conference on Information and Knowledge Management (CIKM)},
  year={2024}
}

@inproceedings{ni25:cikm,
  title={A Content-Driven Micro-Video Recommendation Dataset at Scale},
  author={Y. Ni and Y. Cheng and X. Liu and J. Fu and Y. Li and X. He and Y. Zhang and F. Yuan},
  booktitle={Proceedings of ACM International Conference on Information and Knowledge Management (CIKM)},
  year={2025}
}

@String{Computer = "{IEEE} Computer" }


\end{document}